\documentclass[10pt,twocolumn,letterpaper]{article}

\usepackage{wacv}              

\usepackage{graphicx}
\usepackage{amsmath}
\usepackage{amssymb}
\usepackage{booktabs}

\usepackage{times}
\usepackage{epsfig}
\usepackage{graphicx}
\usepackage{amsmath}
\usepackage{amssymb}
\usepackage{amssymb}
\usepackage{booktabs}
\usepackage{multirow}
\usepackage{pifont}
\usepackage{dsfont}
\usepackage{tabularx}
\usepackage{tabularx}
\usepackage{makecell}
\usepackage[utf8]{inputenc}
\usepackage{enumitem}
\usepackage{adjustbox}

\usepackage{subcaption}
\usepackage[accsupp]{axessibility} 
\usepackage[color=red]{todonotes}
\newcommand{\cmark}{\ding{51}}%
\newcommand{\xmark}{\ding{55}}%

\usepackage[pagebackref,breaklinks,colorlinks]{hyperref}

\usepackage[capitalize]{cleveref}
\crefname{section}{Sec.}{Secs.}
\Crefname{section}{Section}{Sections}
\Crefname{table}{Table}{Tables}
\crefname{table}{Tab.}{Tabs.}

\def\wacvPaperID{2144} 
\def\confName{WACV}
\def\confYear{2024}

\begin{document}

\title{Rank2Tell: A Multimodal Driving Dataset for Joint Importance \\Ranking and Reasoning}  

\author{Enna Sachdeva$^{*1}$, Nakul Agarwal$^{*1}$, Suhas Chundi$^{2}$, Sean Roelofs$^{2}$, Jiachen Li$^{2}$, Mykel Kochenderfer$^{2}$, \\ Chiho Choi$^{1,\dag}$, Behzad Dariush$^{1}$ \\ \small $^{*}$equal contribution  
	     \vspace{.2em}\\
$^{1}$Honda Research Institute USA  \qquad  
$^{2}$Stanford University 
}

\maketitle

\def\thefootnote{\dag}\footnotetext{Now at Samsung Semiconductor Inc.}
\def\thefootnote{1}

\begin{abstract}
  The widespread adoption of commercial autonomous vehicles (AVs) and advanced driver assistance systems (ADAS) may largely depend on their acceptance by society, for which their perceived trustworthiness and interpretability to riders are crucial. In general, this task is challenging because modern autonomous systems software relies heavily on black-box artificial intelligence models. Towards this goal, this paper introduces a novel dataset, Rank2Tell\footnote[1]{https://usa.honda-ri.com/rank2tell}, a multi-modal ego-centric dataset for \textbf{Rank}ing the importance level and \textbf{Tell}ing the reason for the importance. Using various close and open-ended visual question answering, the dataset provides dense annotations of various semantic, spatial, temporal, and relational attributes of various important objects in complex traffic scenarios. The dense annotations and unique attributes of the dataset make it a valuable resource for researchers working on visual scene understanding and related fields. Furthermore, we introduce a joint model for joint importance level ranking and natural language captions generation to benchmark our dataset and demonstrate performance with quantitative evaluations. 
\end{abstract}

\vspace{-0.4cm}
\section{Introduction}
Effective and accurate understanding of visual scenes is an important prerequisite for safe navigation of autonomous vehicles and advanced driver assistance systems, especially in complex and highly crowded urban scenarios. 
Despite significant advancement in the development of self-driving and driver-support technologies, public acceptance of these systems remains limited. A survey conducted by Partners for Automated Vehicle Education (PAVE) in 2020 \cite{Pave} reported that 60\% of respondents would be more likely to trust autonomous vehicles if they have a better understanding of the underlying rationale of the models. To improve the transparency of these systems, intelligent vehicles must have the ability to identify critical traffic agents whose behaviors can influence their own decision making~\cite{zhang2020interaction, ma2021reinforcement, li2021rain, li2022important}. Identifying these important agents allows for a more efficient allocation of computation resources toward predicting the actions of a subset of critical objects and identifying potential risks. 
To establish trust, the autonomous system must provide human-interpretable reasoning about the important agents in the scene, through voice or visual interfaces.

Accurately identifying important agents within the environment and providing human-interpretable reasoning requires a scene understanding model to capture several essential scene features effectively, including 3D mapping, the semantics of the scene, spatial and temporal relations, agents' importance level, actions, intentions, and attention. Additionally, the ability to reason about important agents in a human-interpretable manner is crucial for capturing the essence of the scene. By integrating these comprehensive features, the model's ability to understand and reason about the scene is greatly enhanced. One key to successfully applying these approaches is the availability of traffic datasets with rich human annotations of object importance and reasoning to address the interpretability and trustworthiness of autonomous systems operating in an interacting environment. However, there currently exists no comprehensive real-world driving dataset that provides all these features. 

To address this challenge and facilitate future research, we propose a novel ego-centric, multi-modal dataset for visual scene understanding in urban traffic scenarios. The dataset uses 2D image features and 3D LiDAR features to provide dense semantics, temporal and relational annotations of important agents that influence the ego vehicle's decision making. 
In addition, it provides diverse natural language explanations to enable reasoning about why a particular agent in a scene is of importance. 
We aim to improve the transparency and interpretability of the visual scene understanding modules of autonomous systems. The proposed dataset has the potential to assist drivers in conveying important decisions, improve their situational awareness, and warn passengers about potential safety hazards present in the surrounding environment. We also introduce a model that uses multi-modal 2D+3D features to jointly predict the importance level and generate captions of important agents. 

The contributions of this paper are as follows. \textbf{First}, we introduce the first multi-modal dataset, \textit{Rank2Tell}, for importance level ranking and natural language explanation tasks in urban traffic scenarios. The data is annotated using visual question answering (VQA) that combines video and object-level attributes. \textbf{Second}, we propose a model that uses multi-modal features for joint importance level classification and natural language captioning, and also establish a benchmark suite on these tasks. \textbf{Third}, we introduce the key features of the dataset that can potentially be used to enhance scene understanding for safety-critical applications.

\begin{figure*}[t]
    \centering
    \includegraphics[width=\linewidth]{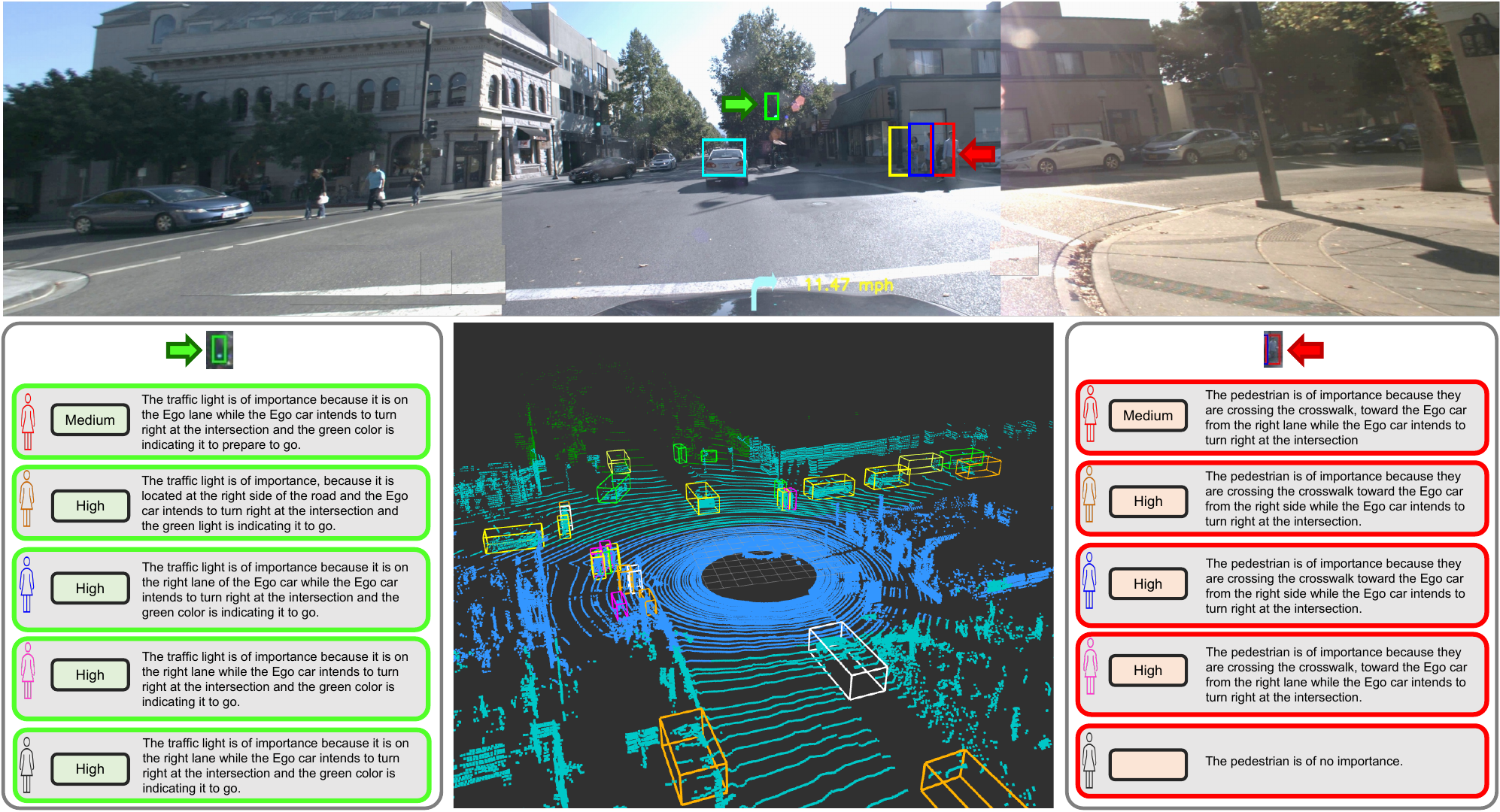}    \vspace{-0.5cm}
    \caption{{\textbf{Overview of Rank2Tell.} 
    Rank2Tell is an ego-centric dataset with three camera images and point cloud features for visual scene understanding in complex urban traffic scenarios.
    For each scenario, five annotators are asked to identify important objects in the scene with three importance levels: high, medium, and low. They also compose natural language descriptions to explain their reasoning behind the importance ranking of each important object in the scene, which leads to diverse annotations of explanations. In this exemplar scenario, the important objects are indicated by bounding boxes. The importance and natural language annotations of the traffic light and the crossing pedestrian are shown as illustrative examples. 
   }} 
    \vspace{-0.2cm}
    \label{fig:main_fig}
\end{figure*}

\section{Related Work}

\subsection{Object Importance Classification}

Identifying important objects in images or videos has attracted a great deal of research interest in various domains. There are several visual saliency studies that have created a pixel-level importance mapping for tasks such as visual question answering \cite{Anderson01, Yu01}, scene understanding \cite{Sun01}, video summarising \cite{Ji01}, driver attention \cite{QLi01}, and AV steering control \cite{Kim01}. However, these methods do not treat objects as distinct instances. Several studies focus on identifying the single most important object in a scene, such as the most important person in a social situation \cite{WLi01}. 
Several other studies make a binary classification between ``important" and ``unimportant" for multiple objects in a scene \cite{Gao02, CLi01, Zhang01, li2022important}. However, only one of these studies \cite{li2022important} attempts to couple natural language with importance estimation but this work only provides a description of a single important object. 
Other studies \cite{wu2022toward, ohn2017all} classify agents in the scenes at three different importance levels: high, medium, and low. However, none of these studies explains the underlying reasoning for the importance estimation for a specific object.

\subsection{Dense Captioning}

There has been a large amount of research on generating natural language to describe visual images or videos. The idea of self-attention was proposed in the visual captioning task \cite{Xu01}, which has been the basis for many further visual captioning research including \cite{Borna01, Devlin01}. Some dense visual captioning methods aim to generate a caption for multiple regions in the same image \cite{Yin01, Yang01, Johnson01} using the Visual Genome dataset \cite{Krishna01}. 
The ScanRefer dataset was proposed for object localization and language description tasks \cite{chen2020scanrefer}. Our problem setting is similar to dense image captioning but we focus on generating explanations for the important objects in traffic scenarios with spatio-temporal observations.

\begin{table*}[!h]
\centering
\footnotesize
\setlength{\tabcolsep}{0.3pt}
\caption{Comparison of our proposed dataset with other datasets.}
\vspace{-0.2cm}
\begin{tabular}{c|cccccc|ccc|ccc}
                      & \multicolumn{6}{c|}{Important Agents Identification only} & \multicolumn{3}{c|}{Captioning only} & \multicolumn{3}{c}{Both} \\ \hline
Dataset               & HDD~\cite{ramanishka2018toward}      & EF~\cite{zeng2017agent}       & SA~\cite{chan2016anticipating}       & DoTA~\cite{yao2020and}   & KITTI~\cite{ohn2017all} & A-SASS~\cite{wu2022toward}   & HAD~\cite{kim2017interpretable}      & BDD-X~\cite{kim2018textual}      & BDD-OIA~\cite{xu2020explainable}      & T2C~\cite{deruyttere2019talk2car} & DRAMA~\cite{malla2023drama}       & Ours       \\ \hline
\# Scenarios &    137    &  3,000      &  1733 &   4,677    & - & 10    & 5,744  & 6,984    & 11,303      &  850 &   17,785    &   116   \\
\# Frames annotated/video    & $\geq$1 & $\geq$1 &  $\geq$1 & $\geq$1 & - & $\geq$1 & $\geq$1 & $\geq$1  & $\geq$1 & 1 & 1 &  $\geq$1 \\

Avg. video duration (in sec) &  2700 &  4 & 5 & - & - & - & 20 & 40 & 
5 & 20 & 2 & 20\\

Importance Localization     &    \cmark       &   \cmark       &     \cmark     &   \cmark        & \xmark   &   \cmark &   \xmark      &      \xmark        &     \xmark           &  \xmark &    \cmark      &   \cmark         \\
Importance Captioning       &     \xmark        &   \xmark    &   \xmark    &    \xmark        &     \xmark    &   \xmark    &     \cmark     &      \cmark      &       \cmark       & \xmark   &    \cmark     &       \cmark     \\
Importance Ranking &    \xmark       &     \xmark       &  \xmark    &    \xmark    &      \cmark      &   \cmark &    \xmark        &    \xmark          &       \xmark         & \xmark &   \xmark        &     \cmark       \\

\# Important objects/frame &    1       &    1   & $\geq$1  &  $\geq$1        &      $\geq$1     &   $\geq$1  & -      &      -      &       -     & - &     1        &     $\geq$1       \\
\# Captions/frame     &     -      &     -   & - &       -   &      -   & - &    1      &     1       &        1    & - &     1        &     $\geq$1       \\
\# Captions/object    &      -     &   -  &   -  &    -      &     -      &  - &  1     &      1      &     1       & - &     1        &     $\geq$1       \\
Avg caption length    &   -    &  -   &  -    &  -     & -      & -       &  11.05 &     8.90    &   6.81         &  11.01  &     17.97        &   31.95   \\
RGB                   &    \cmark       &     \cmark     &      \cmark    &     \cmark  &  \cmark  &  \cmark &  \cmark      &       \cmark     &     \cmark         & \cmark &  \cmark         &   \cmark         \\
LiDAR     &     \xmark        &  \xmark     &  \xmark    &    \xmark       &     \cmark     &  \xmark &  \xmark     &     \xmark     &    \xmark      &  \cmark &  \xmark        &      \cmark      \\
3D boxes                 &      \xmark       &  \xmark   &   \xmark     &    \xmark       &       \cmark   &   \xmark & \xmark    &      \xmark    &      \xmark    & \cmark  &  \xmark        &      \cmark      \\
Field of View (RGB)   &      C     & C  &  C    &    C     &    C     &   LCR & C    &   C       &  C     &  LCR &     C     &   LCR        \\
Object Tracking       &    \xmark       &  \xmark     & \xmark  &     \xmark     &  \cmark   & \xmark     &   \xmark       &      \xmark      &    \xmark      &  \xmark  &    \xmark         &     \cmark       \\
Reasoning             &     \cmark      &    \xmark   & \xmark  &       \xmark   &      \xmark     &   \xmark  &  \xmark    &     \cmark       &     \cmark     &  \xmark  &    \cmark         &    \cmark        \\
Free Form Captions    &    \xmark       &  \xmark     & \xmark  &    \xmark      &     \xmark      &  \xmark &  \cmark      &       \cmark     &    \xmark      &  \cmark  &    \cmark         &    \cmark       
\end{tabular}
\vspace{-0.3cm}
\label{tab:dataset_comp}
\end{table*}

\subsection{Datasets}
In recent years, many traffic scene datasets have been proposed to stimulate progress in the analysis of the important objects in driving scenes. HDD~\cite{ramanishka2018toward}, ROI-A~\cite{agarwal2022risk} and KITTI~\cite{ohn2017all} are object localization datasets that benchmark importance localization and anomaly detection. HAD~\cite{kim2017interpretable}, BDD-X~\cite{kim2018textual}, BDD-OIA~\cite{xu2020explainable} are captioning driving datasets that provide reasons for the ego vehicle's actions in natural language descriptions. 
DRAMA~\cite{malla2023drama} provides important object labels with captions from the ego car's perspective while considering spatio-temporal relationships from videos. While DRAMA~\cite{malla2023drama} is the most relevant dataset to our proposed Rank2Tell, the major differences are shown in Table~\ref{tab:dataset_comp}.

\begin{figure*}[t]
    \centering
    \subfloat[\centering Sequence of Visual Questions asked from annotators during the annotation]{{\includegraphics[width=0.88\linewidth]{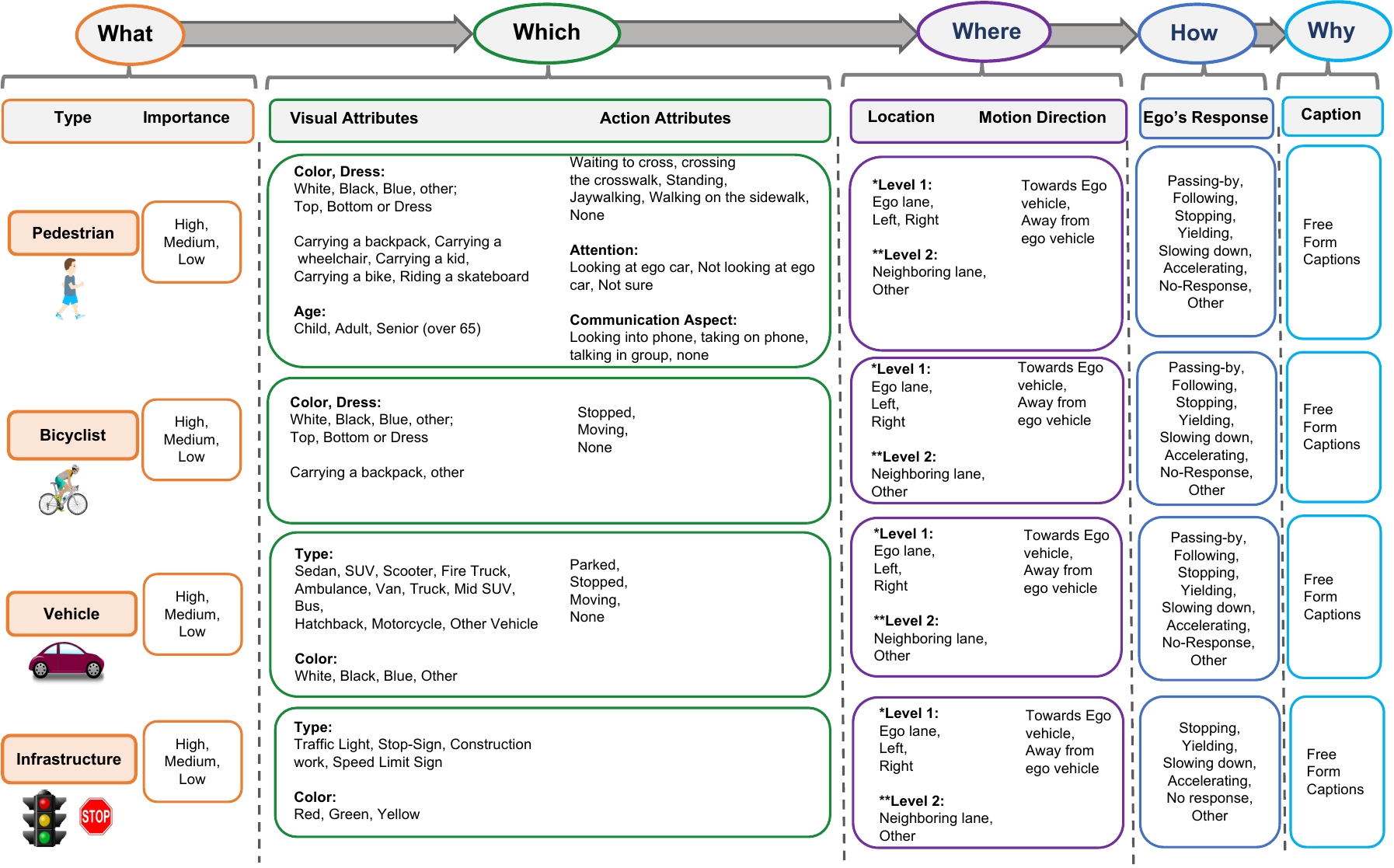} }\label{fig:annotation_ques}}
    \quad
    \subfloat[\centering Location Level  ]{{\includegraphics[width=0.45\linewidth,height=3.5cm]{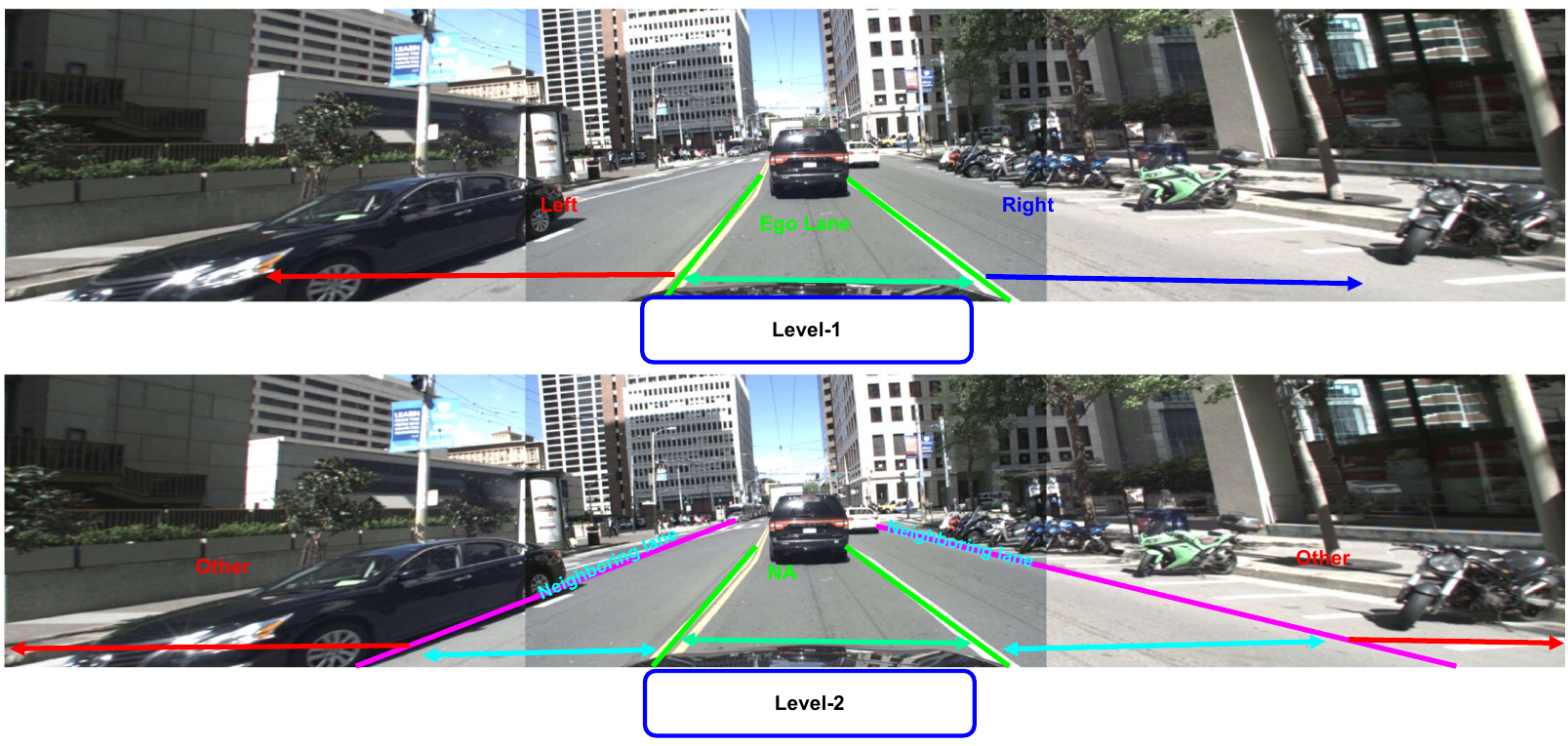} \label{fig:location_level}}}
    \quad
    \subfloat[\centering Motion Direction ]{{\includegraphics[width=0.45\linewidth,height=3.5cm]{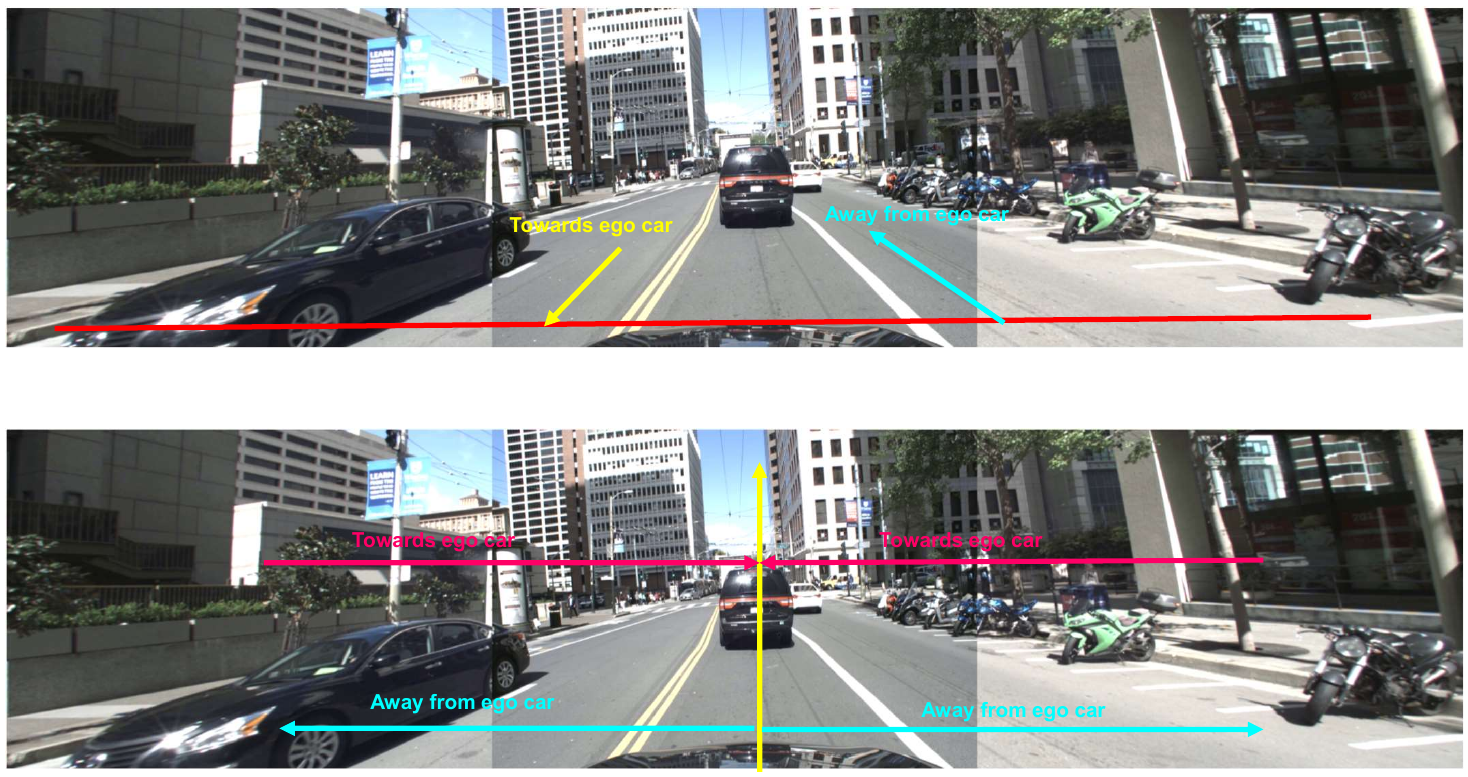} \label{fig:motion_dir}}}
    \vspace{-0.2cm}
    \caption{{The annotation schema of Rank2Tell dataset.}} 
    \vspace{-0.5cm}
    \label{fig:ann_scheme}
\end{figure*}

\begin{figure}[hbt!]
    \centering
    \subfloat[\centering Agents importance level distribution]{{\includegraphics[width=0.9\linewidth,height=4.5cm]{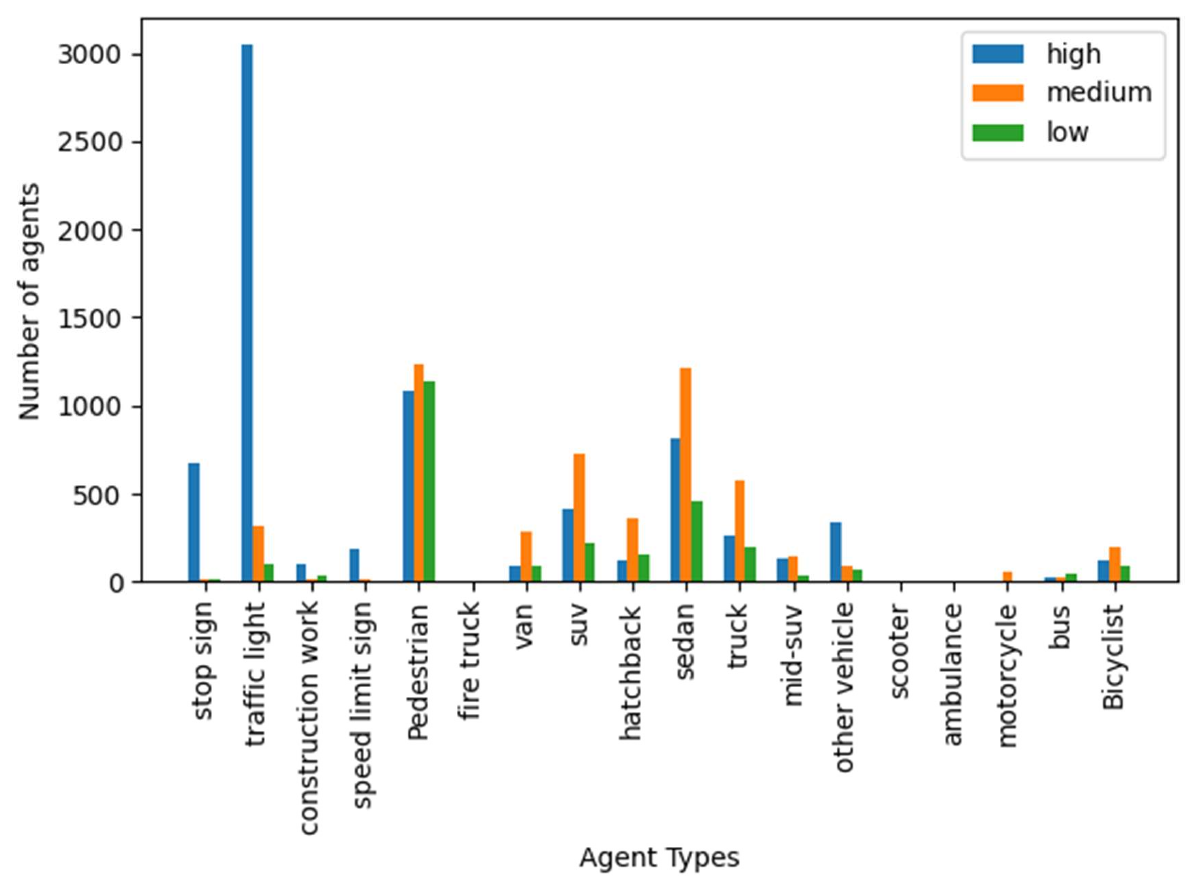}
   }
    \vspace{-0.1cm}
    \label{fig:agents_imp}}
    \\
    \subfloat[\centering Importance level of agents at different locations w.r.t ego's intentions ]{{\includegraphics[width=0.9\linewidth,height = 3.5cm]{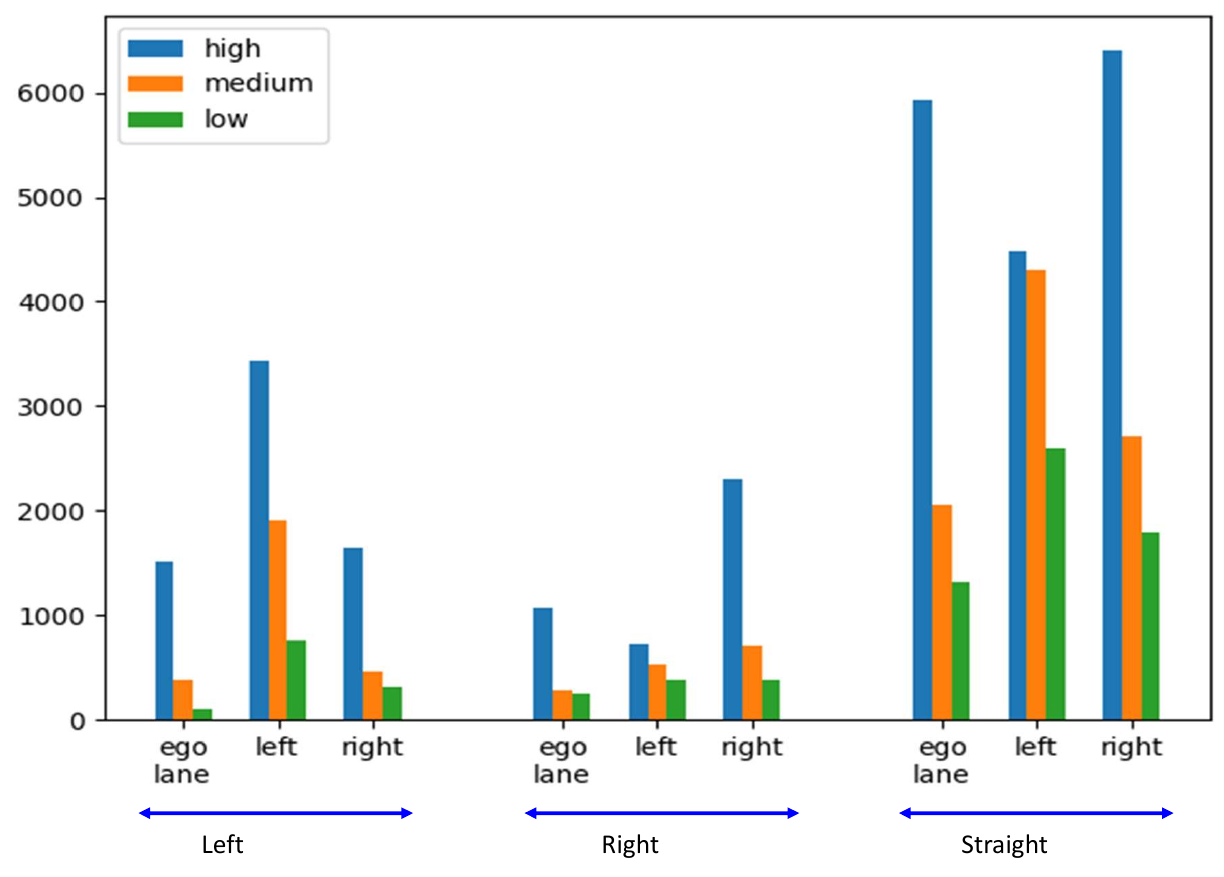} 
    }
   \vspace{-0.1cm}
    \label{fig:intent_loc}
    }
     \caption{Statistical Analysis of Rank2Tell Dataset}
     \vspace{-0.6cm}
    \label{fig:dataset_stats1}
\end{figure}

\begin{figure*}[hbt!]
    \centering
   \subfloat[\centering Top 30 words distribution for captions associated with high important objects]{{\includegraphics[width=0.29\linewidth]{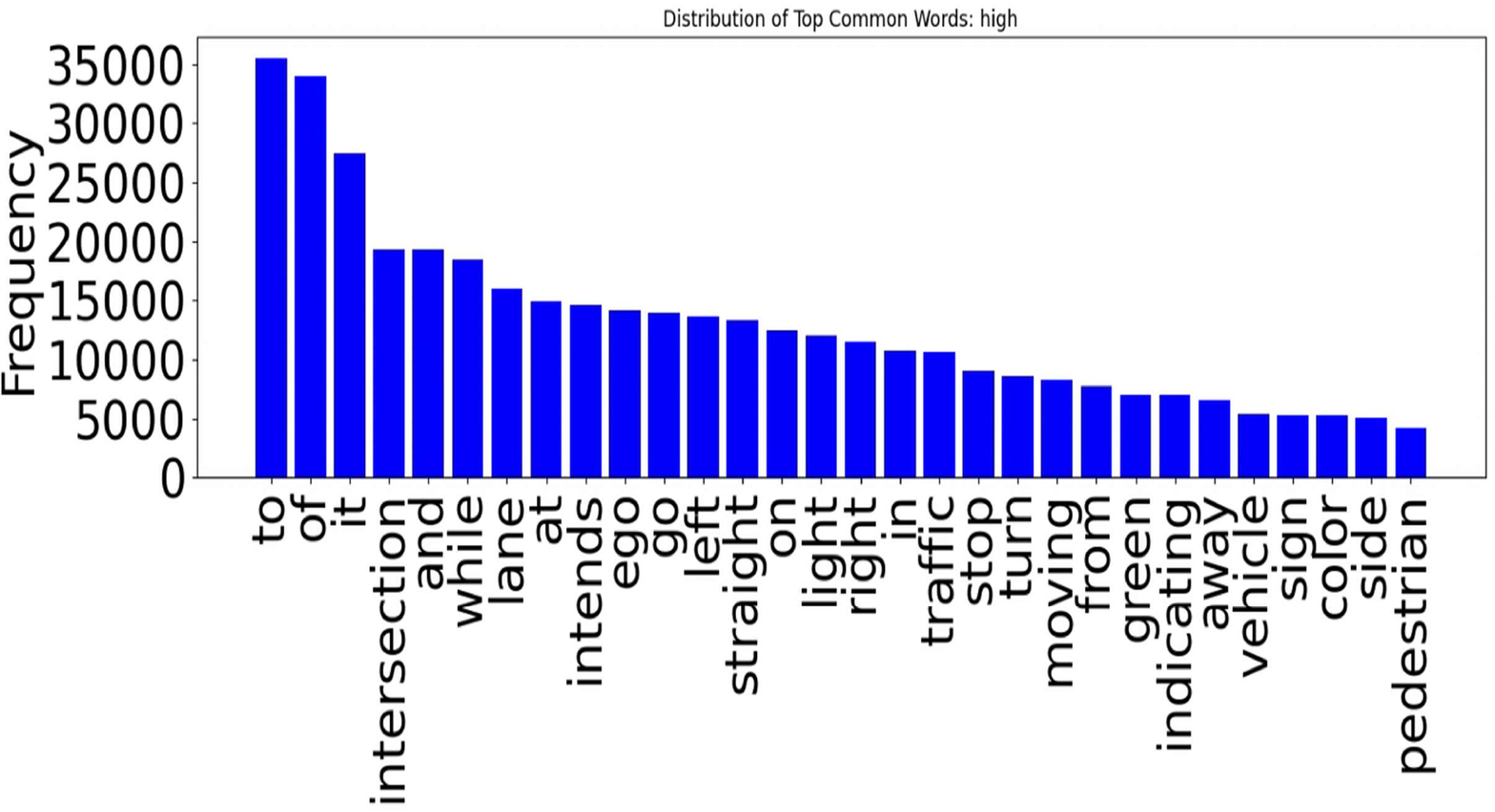}
   }  \vspace{-0.5cm}
    \label{fig:high}}
    \qquad
   \subfloat[\centering Top 30 words distribution for captions associated with medium important objects]{{\includegraphics[width=0.29\linewidth]{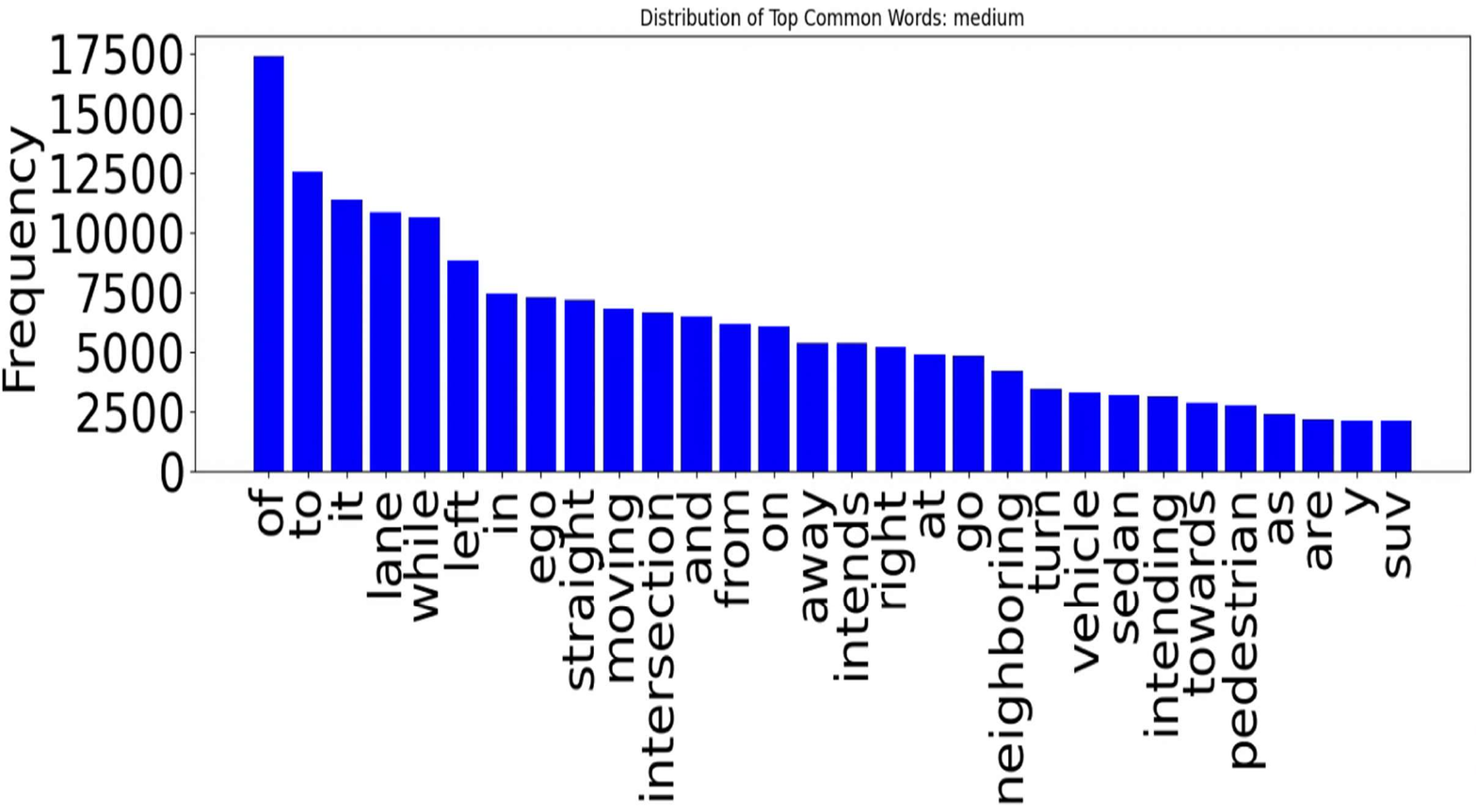}
   } \vspace{-0.5cm}
   \label{fig:medium}}
    \qquad
   \subfloat[\centering Top 30 words distribution for captions associated with high low objects]{{\includegraphics[width=0.29\linewidth]{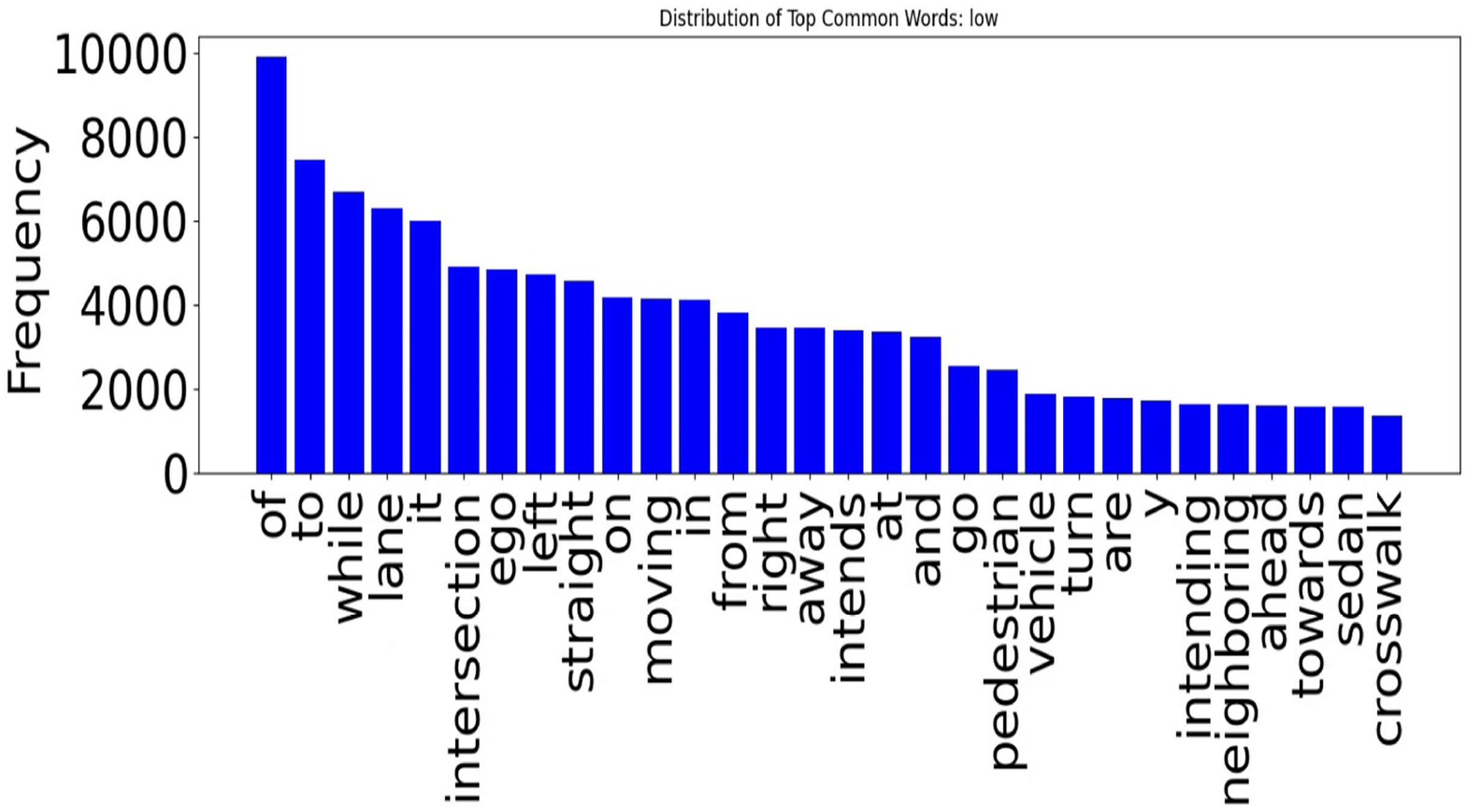}
   } \vspace{-0.5cm}
   \label{fig:low}}
   \caption{Statistical analysis of Rank2Tell dataset.}
    \vspace{-0.5cm}
    \label{fig:dataset_stats}
\end{figure*}

\section{Rank2Tell Dataset}
The dataset is collected using an instrumented vehicle equipped with three Point Grey Grasshopper video cameras with a resolution of $1920\times1200$ pixels, a Velodyne HDL-64E S2 LiDAR sensor, and high precision GPS. Additionally, Vehicle Controller Area Network (CAN) data is collected for analyzing how drivers manipulate steering, breaking, and throttle. All sensor data are synchronized and timestamped using ROS and customized hardware and software. The dataset contains a diverse set of traffic scenes captured at intersections in different environments, including urban areas. We selected 116 clips of approximately 20 seconds each, focusing on intersections from several hours of data. These video clips capture both the entering and exiting of the ego vehicle from the intersection.

\subsection{Annotation Methodology}
The important agent identification is subjective in nature and can vary based on factors such as age, gender, and driving experience. To account for this diversity of opinions, each video was annotated by five annotators with varying levels of driving experience and age. The detailed statistics of annotators are introduced in the supplementary material.

We first stitch images from the three cameras, i.e. front, left, and right, which provide a wide FOV (horizontal 134 degrees). Each frame of a video is overlayed with the ego vehicle's speed, and intention (left/right/straight) while leaving the intersection using an arrow. Annotators annotated every 4th frame of the 10 fps video. 
To avoid bias in important agent identification due to prior knowledge of other agents' future intentions, the annotators had access to only 40 historical frames (i.e., 4 seconds of the video) while annotating a certain frame. 

We divide our annotation scheme into three parts: \textit{Important Agent Identification}, \textit{Important Agent Localization and Ranking}, and \textit{Captioning}. The annotation schema is shown in Figure~\ref{fig:ann_scheme}.

\noindent\textbf{Important Agent Identification.} 
The first step is to identify whether there exist important agents in the scene that have the potential to influence the ego vehicle. 
To this end, we overlay the ego vehicle's intention (i.e., straight, left, right) and speed obtained using CAN bus data, as shown in Figure~\ref{fig:main_fig}.
Then we overlay this information on the stitched images before passing it on to the annotators, which provides important context information to annotators about driver behavior and intent which is crucial to identify important agents. 
This is in contrast with other works in that~\cite{ohn2017all} does not provide this information at all and~\cite{malla2023drama} asks the annotators to label the ego vehicle's intention along with importance, for which they need to watch the entire clip which is not a realistic setting in the real world because the driver does not have access to future information. 
Additionally, while~\cite{malla2023drama} filters raw videos based on the activation of vehicle braking using CAN information, we manually filter data, which is more accurate. The annotators then watch the video and are asked to imagine themselves as the driver and determine if there are important agents present in the scene that may affect their driving. 

\noindent\textbf{Important Agent Localization and Ranking.}
The annotators are instructed to localize each agent in the scene that is important to the ego vehicle by creating a bounding box. At a high level, we asked the annotators to identify agents to which the ego vehicle should be attentive for safe driving. Once an object is marked as important, annotators draw a 2D bounding box around that agent, and rank its importance level and relevance they would have given the object in real driving. 
In this work, we use three levels of importance: \textit{Low}, \textit{Medium}, \textit{High}.

The purpose is to handle ambiguity which is hard to avoid in case of binary categorization, i.e. important or non-important ~\cite{li2022important}, and also to reduce confusion and guesswork which is inevitable in case of a continuous ranking score. 
Inspired by \cite{ohn2017all, wu2022toward}, we posit that two levels of importance (i.e., important and nonimportant) could be overly restrictive and inadequate for handling ambiguous cases. 
Moreover, \cite{wu2022toward} shows that multiple levels could aid drivers’ situational awareness in real time, with the minimal distraction of multiple important objects. Due to the subjectivity of the task, we used five annotators to label each scene to reduce ambiguity and reach some level of consensus. Since different annotators may perceive importance differently, there are cases where the same agent has different levels of importance or is not considered important at all. We show the consistency analysis in Table~\ref{tab:consistency_all}.

\noindent\textbf{Captioning.} 
Our proposed dataset emphasizes explainability as another aspect. In addition to identifying and ranking important agents, we aim to provide an explanation for why these objects are deemed significant. Therefore, after annotators identify and localize the important objects, we request them to annotate certain object-level attributes and utilize them to elaborate on why they regard the object as important via a free-form caption. Specifically, as demonstrated in Fig. \ref{fig:ann_scheme}, we request them to annotate the following:

        \setlist{nolistsep}
    \begin{itemize}[noitemsep]
        \item \textit{What}: What class (type and importance level) does the important agent belong to?
        \item \textit{Which}: Which visual and motion attribute belongs to the agent?
        \item \textit{Where}: Where is the agent (location + direction)?
        \item \textit{How}: How does the ego car respond to the agent?
        \item \textit{Why}: Why is the agent of High/Medium/Low level of importance?
    \end{itemize}

The first four questions require single-choice answers from a pre-defined set of options, while the last question is an open-ended caption that combines the answers to the previous four questions (what, which, where, how). This allows annotators to use free-form captions while incorporating all essential information captured in the 3W+1H format. 

A unique benefit of having multiple captions for a single object is the ability to evaluate caption diversity.  Although caption diversity has been explored in image datasets in previous works~\cite{wang2019describing, dai2017towards, shetty2017speaking}, 
it has rarely been investigated in a video setting, particularly in traffic scenes. These earlier studies concentrated on measuring semantic diversity by various concepts in the same image. In contrast, we focus on diversity based on how humans perceive importance. Due to the subjective nature of this task, it is crucial to assess how different individuals explain and perceive importance. As far as we know, our dataset is the first to provide \textit{diverse} captions for \textit{multiple} objects in traffic scenes.

\subsection{Features and Applications of Rank2Tell}
\noindent\textbf{Scene Graphs.} 
Our proposed dataset introduces several attributes that offer significant potential for generating informative scene graphs by leveraging spatial (i.e., which and where), temporal (i.e., tracking with 2D+3D point cloud features), and semantic (i.e., what and how) features. These relational attributes are captured by scene graphs, using comprehensive scene information recorded by road users. They are valuable for performing downstream tasks.

\noindent\textbf{Situational Awareness.} 
Enhancing situational awareness is critical for safe and efficient navigation in complex traffic scenarios. In a recent study on A-SASS, Wu et al. \cite{wu2022toward} demonstrated the effectiveness of highlighting important agents in the scene to improve drivers' situational awareness. 
Motivated by this study, our proposed dataset can potentially be used to identify important agents in the scene and develop an adaptive user interface for improving the driver's situational awareness in real time. Additionally, our dataset offers unique attributes such as \textit{attention} and \textit{communicative aspects} which can be beneficial in Advanced Driver Assistance Systems (ADAS) applications.

\noindent\textbf{Interpretable Models.} 
Human-interpretable models are important for safety-critical applications like Autonomous Driving or Advanced Driver Assistance Systems (ADAS). The model's interpretability can be evaluated using comprehensive attributes of \textit{Rank2Tell}, and can potentially be used to address several tasks towards providing explanations of driving risks associated with important agents. Some of these tasks are a) \textit{important agent localization and tracking}, b) \textit{importance level ranking}, c) \textit{caption generation}, and d) \textit{diverse captions generation}. Our dataset also enables joint handling of thesetasks, which is another unique aspect. We discuss details of these tasks in supplementary materials.

In this paper, we benchmark the performance on two of these tasks: important agents ranking and captions generation, and also provide a model that jointly addresses them,  as discussed in Section \ref{sec:eval}.

\subsection{Dataset Analysis}
\subsubsection{Dataset Statistics}

Figure~\ref{fig:dataset_stats} shows the distribution of labels obtained using video-level question answering in Rank2Tell. Figure~\ref{fig:agents_imp} shows the distribution of agent types with their importance levels in the scene. This is answered using the \textit{what} in the question answering. Since the dataset comprises all scenarios focused on four-way intersections, the majority of the infrastructures such as traffic lights (3048), and stop signs (668) are of high importance and there is consensus among annotators. Figure~\ref{fig:intent_loc} demonstrates the distribution of location level-1 of various important levels objects given the intention of the ego car. It shows that the majority of agents situated on the left and right of the ego car are of high importance when the ego car's intention is left and right, respectively. However, when the ego car's intention is to go straight, the agents' location is equally distributed in three lanes- left, right, and ego lane. This makes the task of importance ranking classification nontrivial and difficult to estimate by only using the ego car's intention as a feature.

To describe the visual attributes of significant objects, annotators usually provide free-form responses that integrate information about the object's what, where, how, and which. 
In Figure~\ref{fig:dataset_stats}, we illustrate the distribution of the top 30 words utilized in captioning the ``why'' question. This demonstrates that annotators effectively conveyed the intention, motion direction, and location of important agents while generating natural language captions.

\vspace{-0.3cm}
\subsubsection{Consistency Analysis}
We conduct an inter-annotator consistency analysis among five annotators, based on the mode of various importance levels they selected, as presented in Table \ref{tab:consistency_all}. In cases with multiple modes, we set the highest importance level as the final importance to obtain these consistency scores. A $40\%$, $60\%$, $80\%$, and $100.00\%$ consistency implies that 2, 3, 4, and 5 annotators provided the same importance levels for an object. Results show that for data instances with \textit{High} as the majority importance, $88.23\%$ of data samples exhibit more than $60\%$ consistency. Similarly, for data instances with \textit{Not-Important} as the majority voting, $98.61\%$ data samples have more than $60\%$ consistency. This indicates that annotators highly agreed on objects selected as \textit{High} and \textit{Not-Important} importance based on mode. 
To assess the quality of the dataset, we compute intra-class correlation (ICC) \cite{shrout1979intraclass}, which is widely used~\cite{rasouli2019pie, agarwal2023ordered} for the assessment of consistency made by different observers measuring the same quantity. The ICC for our annotations is 0.92, which shows excellent inter-rater agreement \cite{cicchetti1994guidelines}. 
However, we use the mode of only importance level classes to obtain the ground-truth importance level of an object for various tasks, such as object importance level classification (Section \ref{section:importance_baseline}). That is, if 2 out of 5 annotators deem an agent as important, we use the ground truth as the mode of the two importance levels instead of all 5. This method aims to reduce falsely underestimating an agent’s importance. Please refer to the supplementary materials for more details.

\begin{table}[h]
\centering
\small
\caption{Data (\%) for different inter-annotator consistency (\%) of agent's importance annotation based on the mode of all 5 annotators' importance levels.}
\vspace{-0.2cm}
\begin{tabular}{c|c|c|c|c}
                
Consistency & Not Important & Low & Medium  & High \\ 

    \hline
    
    40 &  1.37 & 38.33 & 33.77 & 11.75 \\
    
    60 & 4.52 & 51.54 & 42.39 & 30.93 \\

    80 &  7.86 & 9.46 & 17.79 & 28.95\\
    
    100 & 86.23 & 0.65 & 6.03 & 28.35\\

    \hline
    $\geq$60 &  98.61 & 61.65 & 66.21 & 88.23\\

\end{tabular}
\vspace{-0.5cm}
\label{tab:consistency_all}
\end{table}

\begin{figure*}[t]
    \centering
    \includegraphics[width=\linewidth]
    {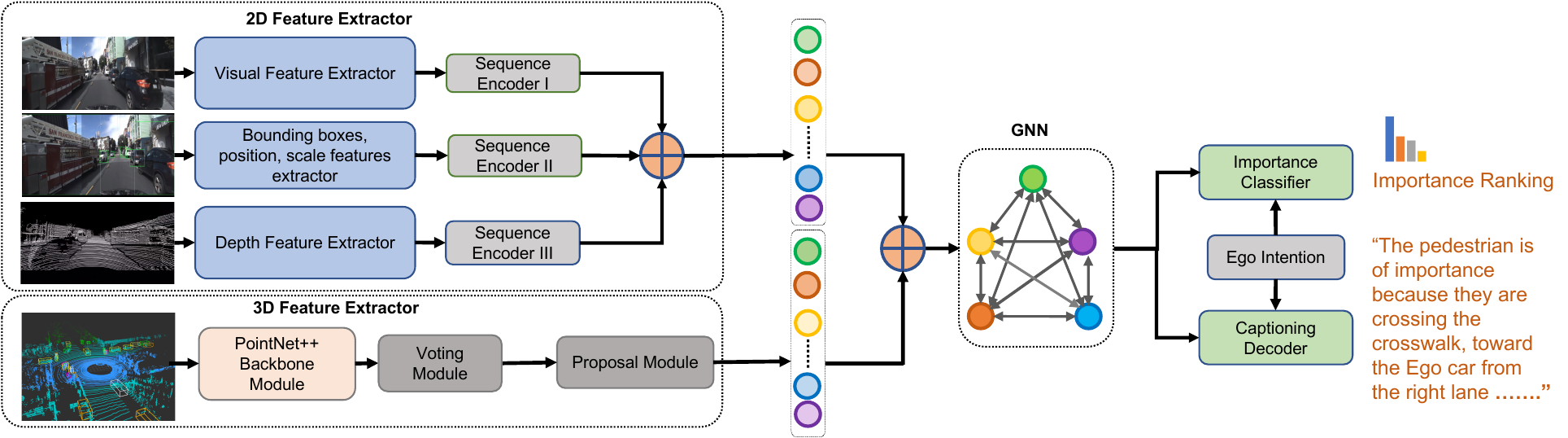}    
    \vspace{-0.5cm}
    \caption{\textbf{Architecture of Joint Model.} The framework consists of five components: a 2D feature extractor, a 3D feature extractor, a relational graph, an importance classifier, and a captioning decoder.}
    \vspace{-0.3cm}
    \label{fig:joint_model}
\end{figure*}

\section{Methodology}\label{sec:method}
We introduce a model to jointly address objects' importance classification and caption generation.
The architecture of the model is illustrated in Figure \ref{fig:joint_model}, which leverages multiple modalities (i.e., 2D and 3D features). It consists of five main components: a) a 2D deep feature extraction module that extracts object-level features from frontal-view visual observations and ego vehicle state information, b) a 3D deep feature extractor that extracts object-level features from 3D point cloud observations, c) a relational graph module that learns the enhanced objects (node) features and relational (edge) features, d) an importance classification module that takes in the objects and objects relational features to predict the importance level for each important objects in the scene, and e) a context-aware attention captioning module that generates descriptions from the object features and relational features. 

\subsection{Feature Extractor}\label{sec:feature_extrac}

We adopt a 2D deep feature extractor from \cite{li2022important}. The deep feature extractor uses a sequence of RGB images, depth images, semantic maps, the ego vehicle's inertial data, and a 2D bounding box of each object in the scene. The depth images are obtained by projecting the point cloud to the stitched camera view and the segmentation maps are obtained using DeepLabv3 on the stitched RGB images. 
To extract the 3D features of each object, we adapt the Pointnet++ \cite{qi2017pointnet++} backbone and the voting module in VoteNet \cite{ding2019votenet} to aggregate all object candidates in a scene to individual clusters. 
To capture the mutual influence and relations among objects, we use a graph-based approach that models objects as nodes and their relations as edges in a relational graph module. The module takes in the concatenated 2D and 3D object features and extracts both object features and relational features between objects. 
To model the relational (edge) features, the module considers only the $K$ nearest objects surrounding the target object to limit the computation complexity. The final object features are obtained by concatenating the graph node features with the object relations, global features, ego features, and ego intentions. These are then fed into the importance classifier and the captioning decoder, respectively. More details can be found in the supplementary materials.

\subsection{Training Objective}

The loss function for our joint model consists of importance classification loss and caption generation loss. We apply a conventional cross-entropy loss on the generated token probabilities, as in previous works \cite{li2022important, chen2021scan2cap}. We used a weighted sum of both loss terms as the final loss:
\begin{equation}
\mathcal{L} = \alpha \mathcal{L}_\text{imp} + \beta \mathcal{L}_\text{cap},
\end{equation}
where $\alpha$ and $\beta$ are the weights for the loss terms.
Further, to enforce the model to reduce the instances of falsely underestimating an agent’s importance, we penalize the $\mathcal{L}_{imp}$ corresponding to different ground truth (GT) and predictions (P) for different importance levels of high (H), medium(M), low(L), and not-importance(NI) as follows:
\begin{align}
\mathcal{L}_{imp}  & = \Sigma_{i=1}^{N} \mathcal{L}_{i} \\
\mathcal{L}_{i} & =\begin{cases}
    \lambda_k \mathcal{L}_{i}, & \text{if $(GT - P) = k > 0$},\\
    \mathcal{L}_{i} , & \text{otherwise},
  \end{cases}
\end{align}
where $\mathcal{L}_{i}$ is the cross-entropy loss for each object $i$. 

\section{Experiments}\label{sec:eval}
We evaluate the performance of our model on the proposed Rank2Tell dataset by comparing it with various baselines on two tasks: importance level ranking (classification) and natural language captions generation. For fair comparisons, we use the same backbone modules and hyperparameters across different baselines. More details can be found in the supplementary materials.

\begin{table}[!tbp]
\centering
\begin{tabular}{c|ccc} 
Method       & F1 (I) & F1 (NI) & Accuracy \\ \hline
OIE\cite{gao2019goal} & 55.78 &  87.74 & 80.80 \\
INTERACT\cite{zhang2020interaction} & 56.42 & 88.02 & 81.21 \\
IOI\cite{li2022important}  & 64.01  &  89.06 & 83.22  \\ \hline 
Ours & \textbf{78.44} & \textbf{92.97} & \textbf{89.39} \\ 
\end{tabular}
\caption{Quantitative evaluation of F1 scores for two importance levels across baselines (I: important, NI: nonimportant).}
\vspace{-0.3cm}
\label{table:binary}
\end{table}

\subsection{Importance Level Classification: Baselines}
\label{section:importance_baseline}

All the importance classification baselines take 2D image features and object features as inputs and infer the importance level of objects in the scene.

\noindent\textbf{Goal-oriented object importance estimation (OIE)\cite{gao2019goal}:}  OIE is a two-stage framework that firstly generates object tracklets from videos as object proposals and then classifies the proposals as important objects.

\noindent\textbf{Interaction graphs based object importance estimation (INTERACT)\cite{zhang2020interaction}:} INTERACT is a graph convolutional network based method that extracts appearance features from objects and represents them as nodes in the graph. These features are then updated by interacting with other nodes through graph convolution based on the learned interaction graph edges. The updated features are used to estimate object importance using a multi-layer perceptron. 

\noindent\textbf{Important Object Identification (IOI)\cite{li2022important}: } 
IOI is a graph convolutional network based method that explicitly models both the appearance and motion of objects in the scene and also uses the ego vehicle state information in the current frame for importance classification. It reasons about the relations between objects in the scene.

\begin{table}[!tbp]
\setlength{\tabcolsep}{1.8pt}
\begin{tabular}{c|ccccc} 
Method       & F1 (L) & F1 (M)& F1 (H) & F1 (NI) & Accuracy \\ \hline
OIE\cite{gao2019goal} & 10.25 &  14.32 & 49.68 & 86.56 & 74.95\\
INTERACT\cite{zhang2020interaction} & 14.04 & 27.56 & 44.04 & 87.45 & 75.09\\
IOI\cite{li2022important}  & 14.09 & 16.19 & 49.55 & 87.85 & 76.84  \\ \hline
Ours & \textbf{20.49} & \textbf{28.94} & \textbf{58.84} & \textbf{92.62} & \textbf{80.93} \\
\end{tabular}
\caption{Quantitative evaluation comparing the F1 scores for 4 importance levels across baselines. L:LOW, M:MEDIUM, H:HIGH, NI:NON-IMPORTANT}
\label{table:importance}
\end{table}

\begin{table}[!tbp]
\label{binary_class}
\centering
\begin{tabular}{c|cccc} 
Method    &  C & B-4 & M & R \\ \hline
S$\&$T~\cite{vinyals2015show}  & 47.67 & 30.24 & 34.91 & 53.31\\
Scan2Cap~\cite{chen2021scan2cap} & 56.32 & \textbf{49.59} & \textbf{38.36} & 66.35 \\ \hline
Ours & \textbf{100.15} & 45.83 & 36.21 & \textbf{68.56}\\
\end{tabular}
\caption{Quantitative evaluation comparing the performance of baselines for captions predictions. C: CIDER, B-4: Bleu-4, M: Meteor, R: Rouge}
\vspace{-0.3cm}
\label{table:caption}
\end{table}

\vspace{-0.1cm}
\subsection{Captioning: Baselines}
\noindent\textbf{Scan2Cap \cite{chen2021scan2cap}:} Scan2Cap is an end-to-end method to perform dense captioning on 3D point clouds to densely detect and describe 3D objects in RGB-D scans. The method first employs a detection pipeline to obtain object proposals and then applies a relational graph and context-aware attention captioning module to learn object relations and generate tokens, respectively.       

\noindent\textbf{Show, and Tell  (S$\&$T)\cite{vinyals2015show}: }This baseline generates captions using 2D features (i.e., global RGB image features and target object features). The visual features are extracted using a ResNet-101 pre-trained on the ImageNet dataset. The global features are concatenated with the target object features, which are used to generate the captions. 

\subsection{Metrics}
To measure the quality of the predicted importance ranking of each object in the scene and their corresponding captions, we evaluate the performance of the baselines using standard classification metrics: F1 score, and accuracy for the importance estimation, and standard metrics such as BLEU-4 (B4) \cite{papineni2002bleu}, METEOR (M) \cite{banerjee2005meteor}, ROGUE (R) \cite{lin2004rouge}, and CIDER (C) \cite{vedantam2015cider}. 

\subsection{Results}

\subsubsection{Quantitative Comparison}
Table \ref{table:binary}, Table \ref{table:importance}, and Table \ref{table:caption} show the quantitative results of our model and baselines for importance classification and caption prediction tasks. The importance baselines are trained and evaluated with two classes (Important and Non-Important), and four classes (High, Medium, Low, and Non-Important). For the two classes, high, medium, and low important objects are merged as one important class. 
The INTERACT \cite{zhang2020interaction} baseline outperforms OIE \cite{gao2019goal} as it models agent interactions in the scene using graph convolutional networks. This makes it more efficient in predicting the importance levels of multiple agents in the scene with lesser input features than OIE \cite{gao2019goal}. 
Furthermore, IOI\cite{li2022important} outperforms the previous two baselines as it utilizes the ego vehicle's speed, acceleration, and intention features. While the agent's importance level is highly influenced by the ego vehicle's intention (e.g., straight, left, right) at the intersection, the ego's speed and acceleration are influenced by the importance level of an agent. This interaction behavior is explicitly incorporated in IOI~\cite{li2022important}, which makes it outperform other methods. Since there is currently no importance ranking estimation model that uses 3D features, we benchmark results on baselines using only 2D features.  

For the captioning baselines, Scan2cap\cite{chen2021scan2cap} outperforms S$\&$T\cite{vinyals2015show} because Scan2Cap uses 3D point clouds and 3D bounding box features, which are more informative than 2D features. The 3D point cloud provides information on the distance between agents and their heights, which captures the dependence of the agents' location with the words in the captions, such as \textit{left side} or \textit{right side} of the ego car.
Our proposed joint model leverages both 2D and 3D features and outperforms the uni-modal baselines in several metrics. Further, the joint training of importance classification and captions prediction complements each other, thereby outperforming the baselines trained with a single task (more details in supplementary material).

To demonstrate the usefulness of Rank2Tell annotations, we conduct an ablation study where we integrate action attributes ($Which$) with object features for all baselines, and the detailed results are shown in supplementary materials.

\section{Conclusion}
We present a novel multi-modal dataset (Rank2Tell) in urban traffic scenarios, which enables joint importance level prediction and reasoning in traffic scenes with frame-level annotations. The dataset includes object-level questions on important objects and language interpretations of the scene from the ego driver's perspective, integrating spatial, temporal, and relational features. This offers new avenues for improving visual scene comprehension and advancing autonomous systems' interpretability and trustworthiness.

{\small
\bibliographystyle{ieee_fullname}
\bibliography{egbib}
}

\end{document}


\title{Rank2Tell: A Multimodal Driving Dataset for Joint Importance \\Ranking and Reasoning: Supplementary Materials}  

\author{Enna Sachdeva$^{*1}$, Nakul Agarwal$^{*1}$, Suhas Chundi$^{2}$, Sean Roelofs$^{2}$, Jiachen Li$^{2}$, Mykel Kochenderfer$^{2}$, \\ Chiho Choi$^{1,\dag}$, and Behzad Dariush$^{1}$ \\ \small $^{*}$equal contribution  
	     \vspace{.2em}\\
$^{1}$Honda Research Institute USA \qquad
$^{2}$Stanford University 
}

\maketitle
\def\thefootnote{\dag}\footnotetext{Now at Samsung Semiconductor Inc.} 


\maketitle
\section{Rank2Tell Dataset}
\subsection{Example Scenarios}


\begin{figure*}[!htp]
    \centering
  \begin{subfigure}[b]{\linewidth}
    \includegraphics[width=0.9\linewidth, height=7.5cm]{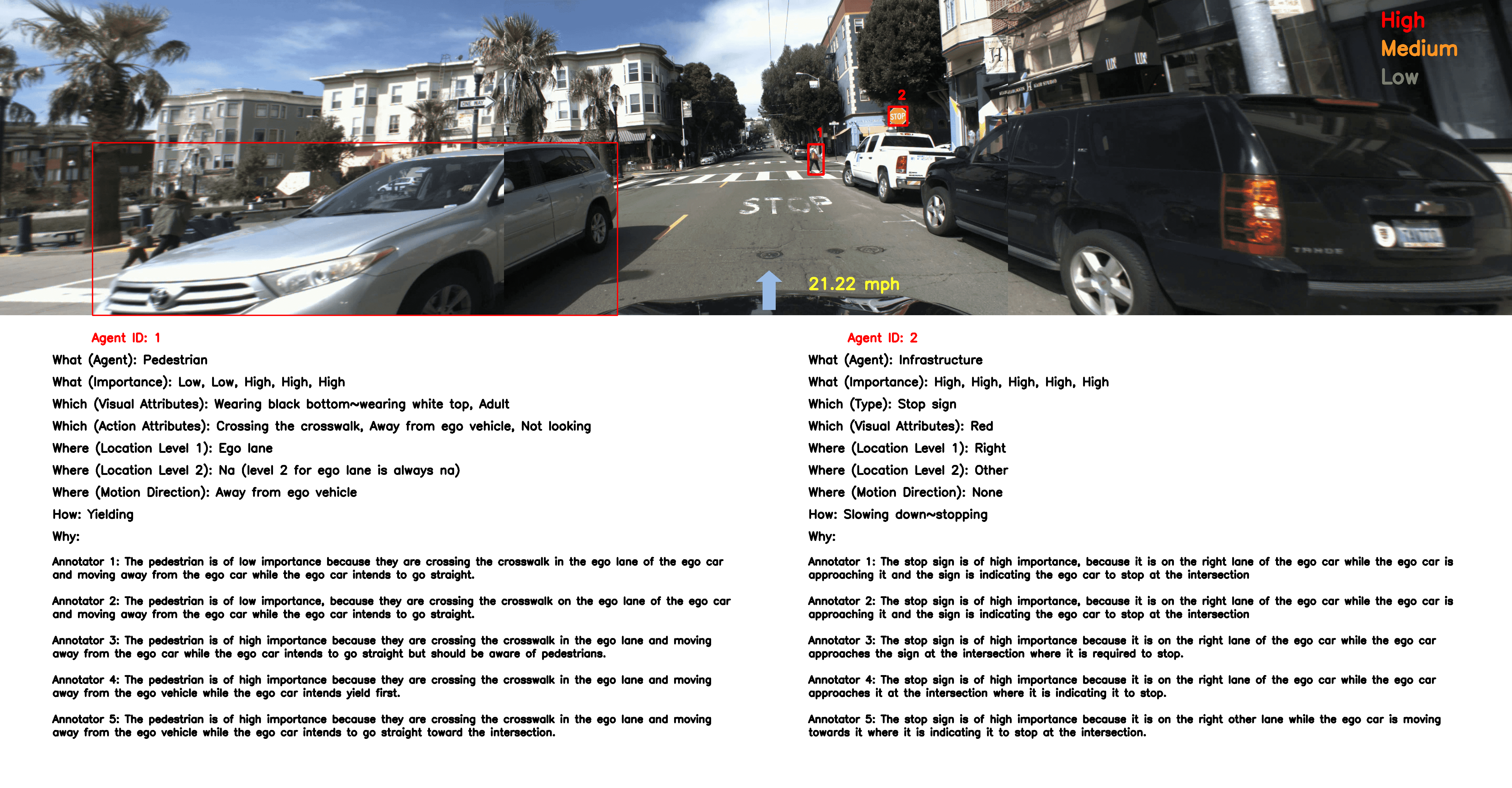}
    \caption{}
    \label{fig:straight_intent_1}
  \end{subfigure}
   \hfill
  \begin{subfigure}[b]{\linewidth}
    \includegraphics[width=0.9\linewidth, height=7.5cm]{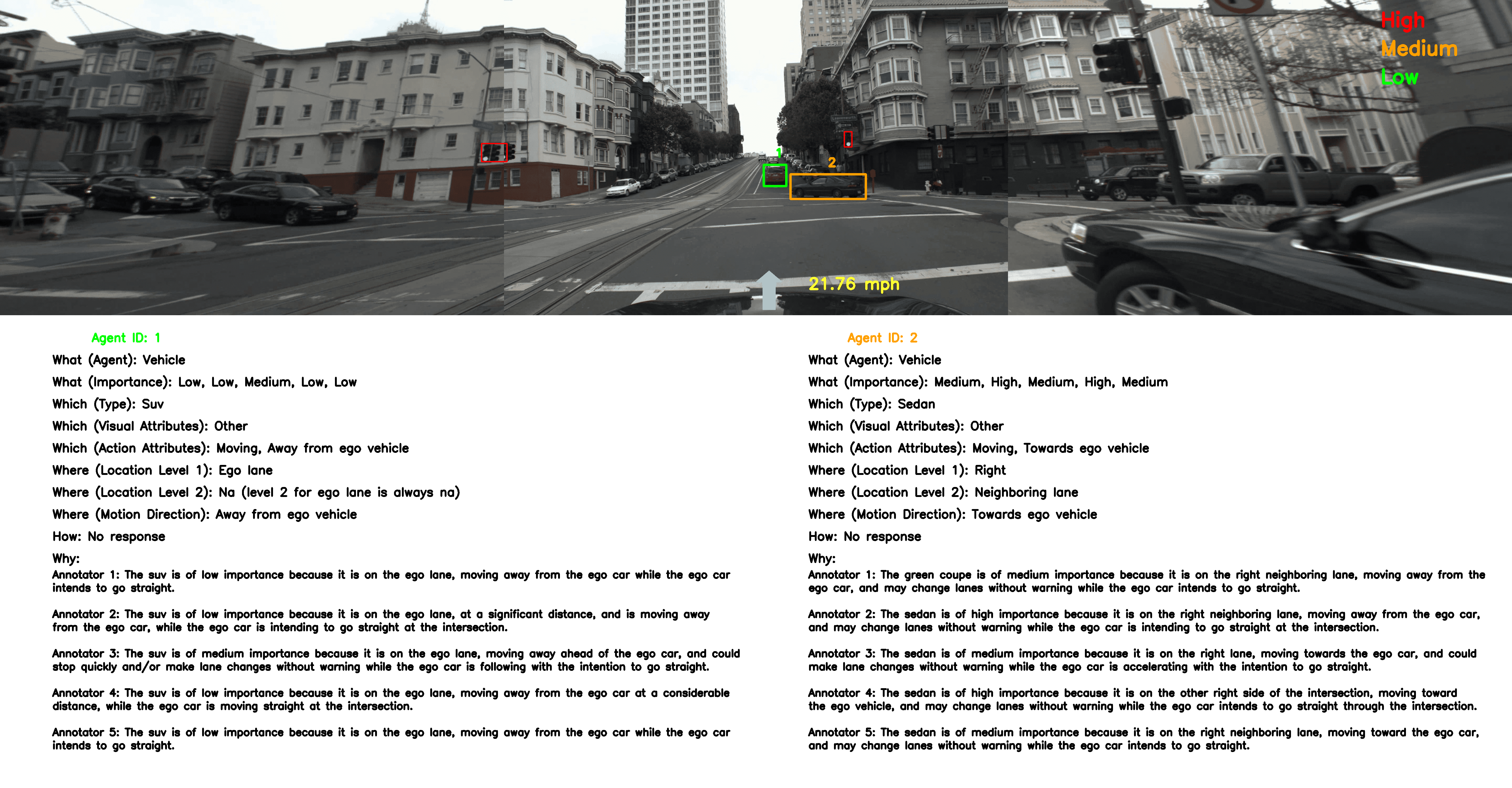}
    \caption{}
    \label{fig:straight_intent_2}
  \end{subfigure}
   \hfill
  \begin{subfigure}[b]{\linewidth}
    \includegraphics[width=0.9\linewidth, height=7.5cm]{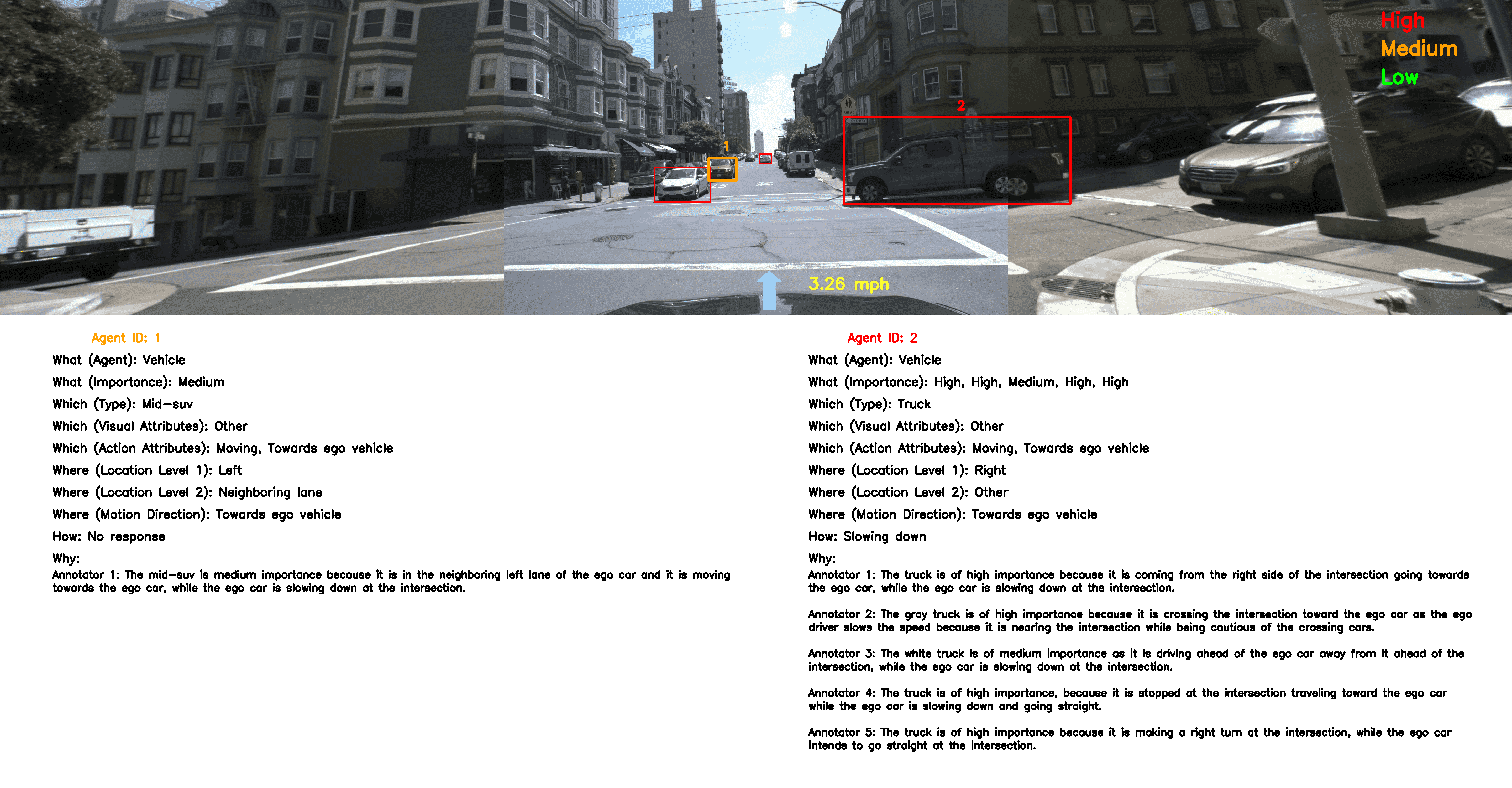} 
    \caption{}
    \label{fig:straight_intent_3}
  \end{subfigure}
    \caption{{\textbf{Example scenarios of Rank2Tell with ego intention to go straight at the intersection}}} 
    \label{fig:example_scenarios_straight}
\end{figure*}


\begin{figure*}[!htp]
    \centering
  \begin{subfigure}[b]{\linewidth}
    \includegraphics[width=0.9\linewidth, height=7.5cm]{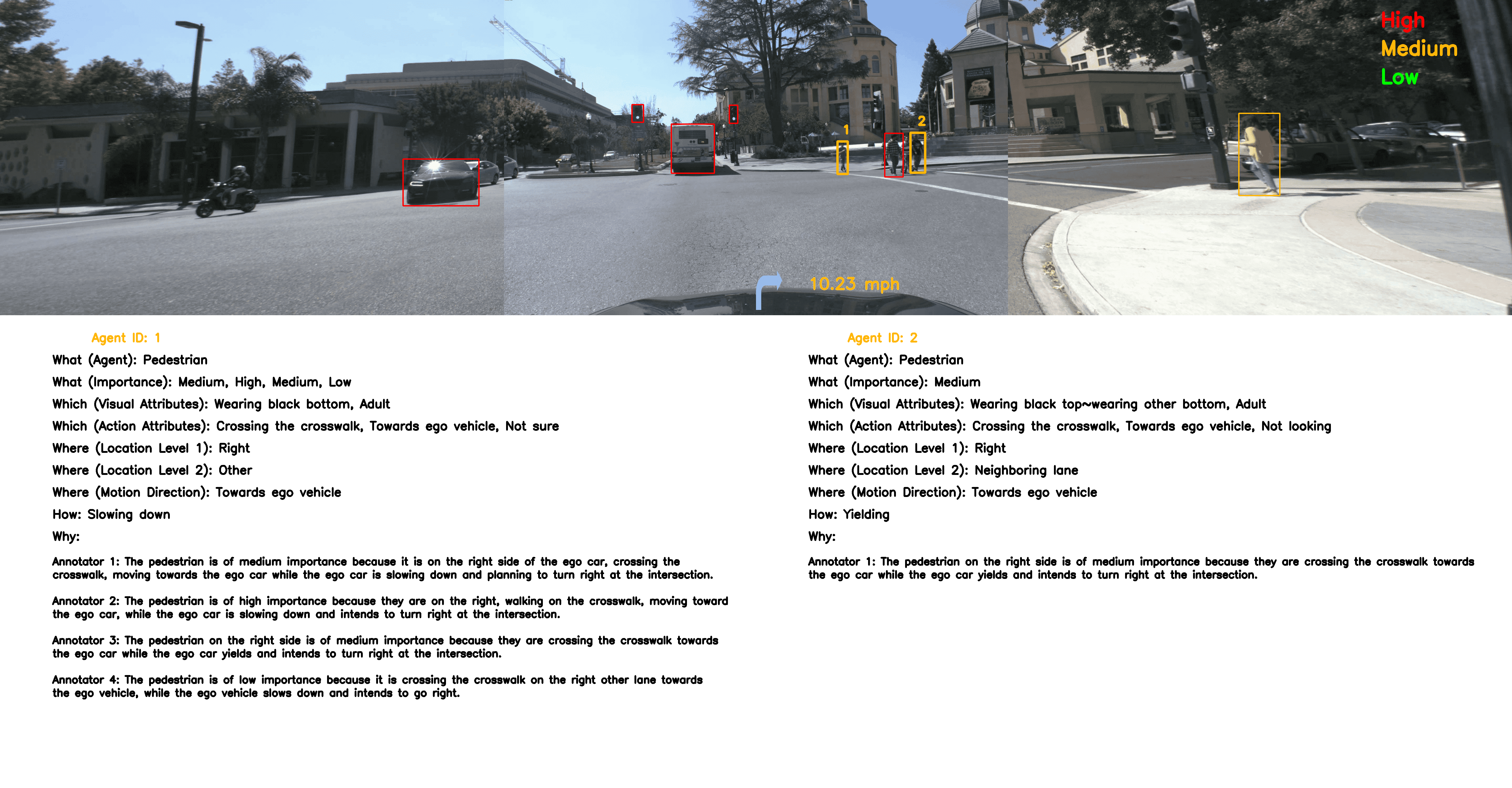}
    \caption{}
    \label{fig:right_intent_1}
  \end{subfigure}
   \hfill
  \begin{subfigure}[b]{\linewidth}
    \includegraphics[width=0.9\linewidth, height=7.5cm]{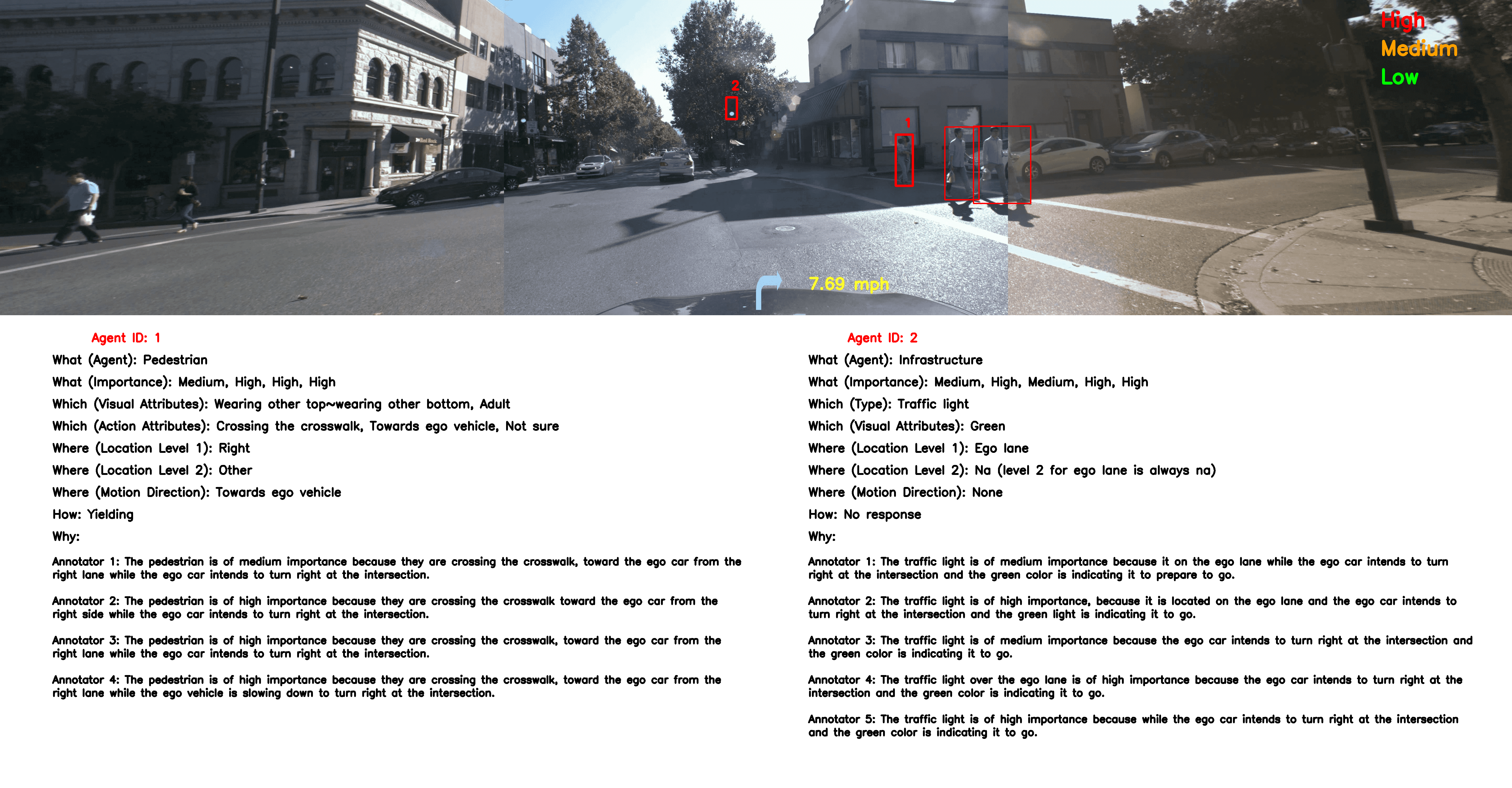}
    \caption{}
    \label{fig:right_intent_2}
  \end{subfigure}
   \hfill
  \begin{subfigure}[b]{\linewidth}
    \includegraphics[width=0.9\linewidth, height=7.5cm]{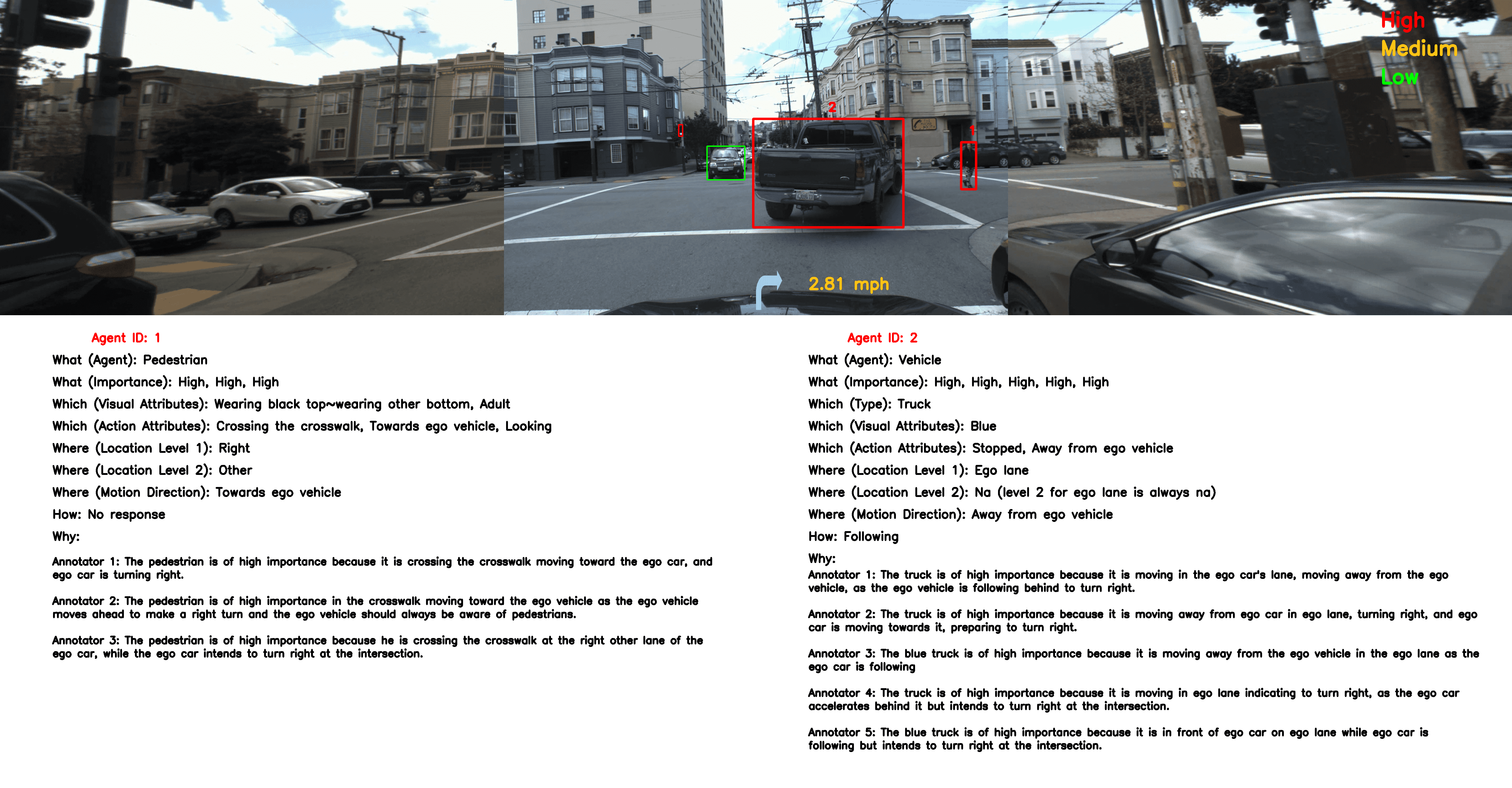}
    \caption{}
     \label{fig:right_intent_3}
  \end{subfigure}
    \caption{{\textbf{Example scenarios of Rank2Tell with ego intention to turn right at the intersection}}} 
    \label{fig:example_scenarios_right}
\end{figure*}

   

\begin{figure*}[!htp]
    \centering
  \begin{subfigure}[b]{\linewidth}
    \includegraphics[width=0.9\linewidth, height=7.5cm]{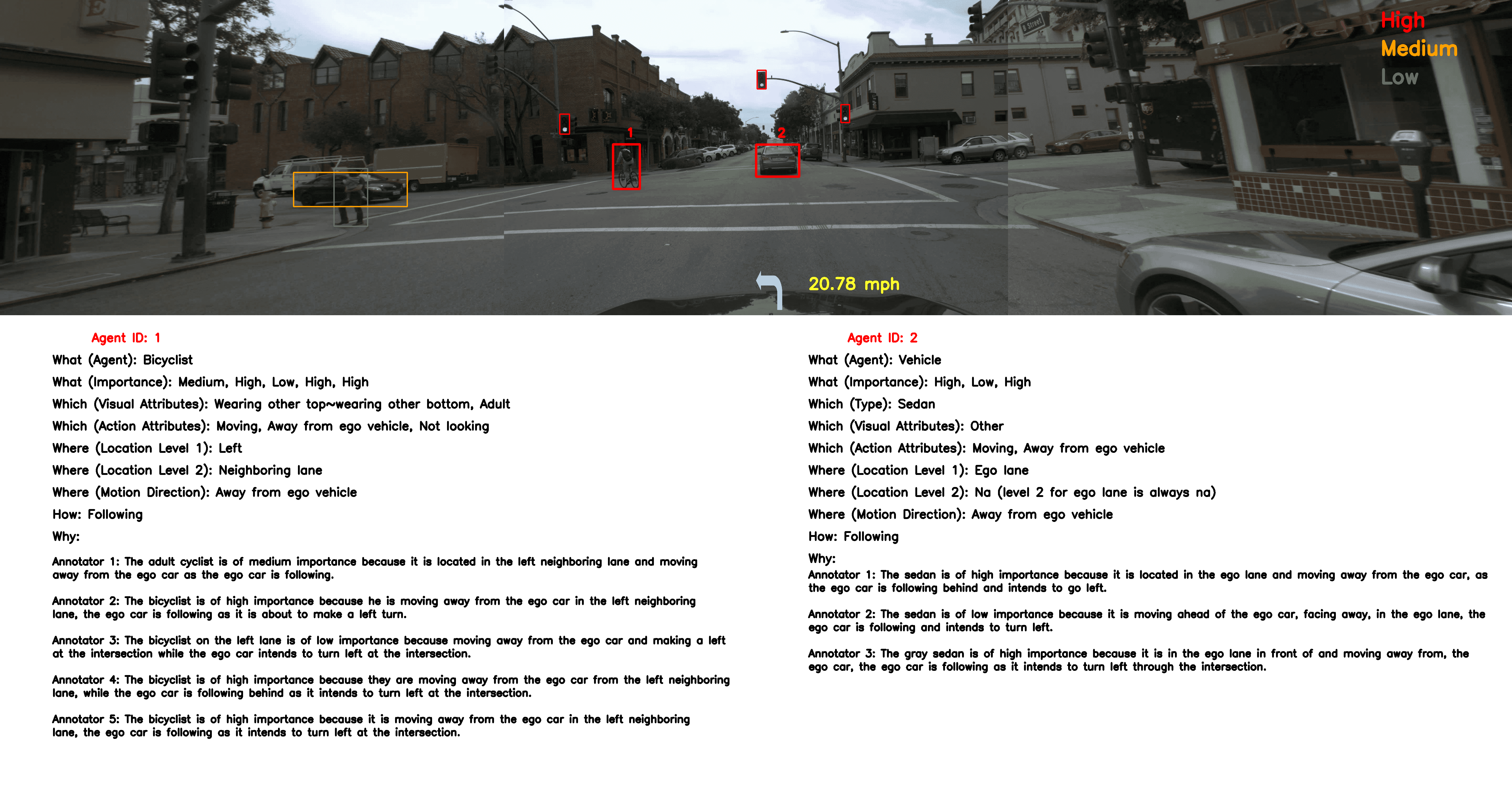}
    \caption{}
    \label{fig:left_intent_1}
  \end{subfigure}
   \hfill
  \begin{subfigure}[b]{\linewidth}
    \includegraphics[width=0.9\linewidth, height=7.5cm]{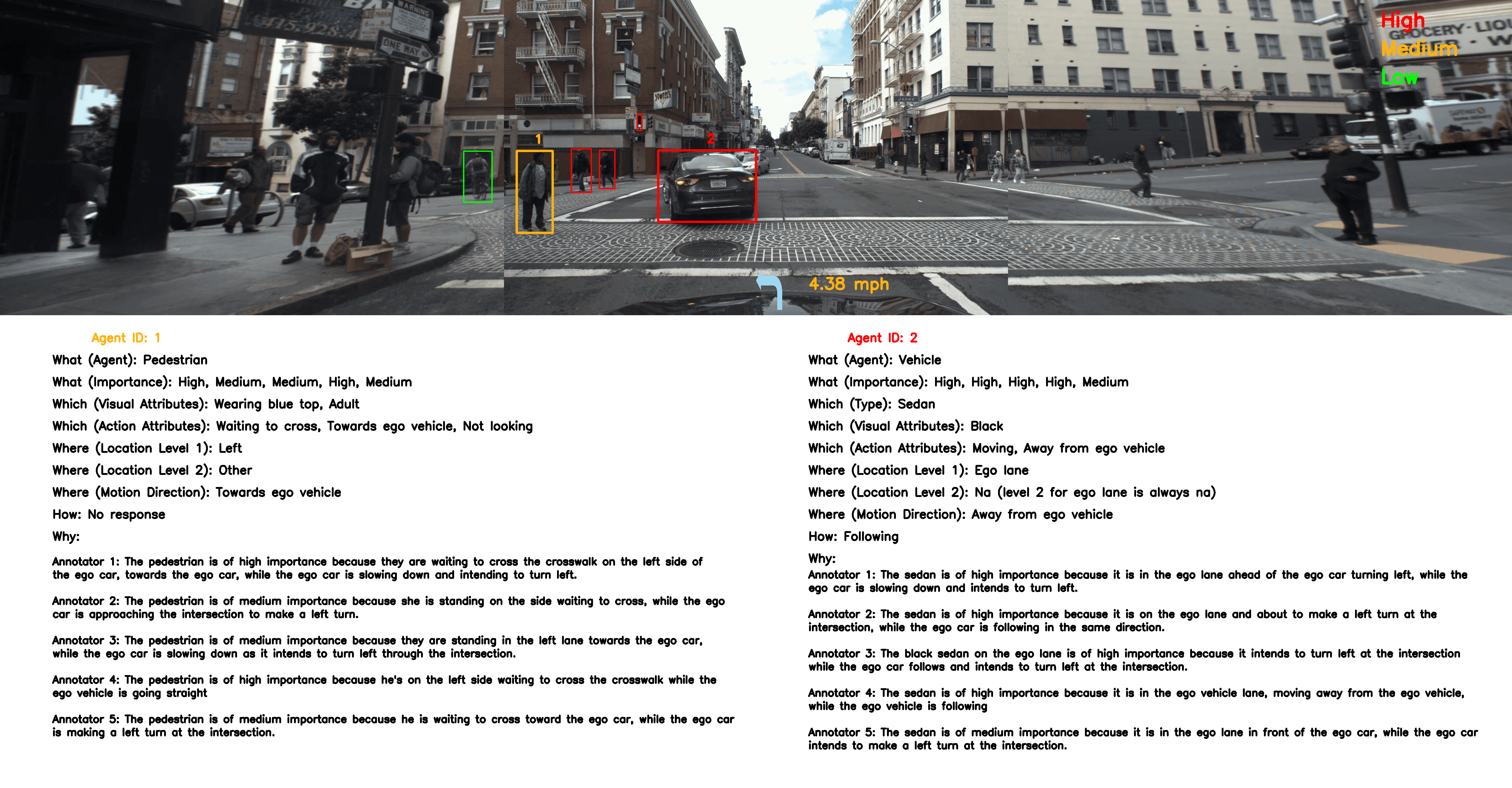}
    \caption{}
    \label{fig:left_intent_2}
  \end{subfigure}
   \hfill
  \begin{subfigure}[b]{\linewidth}
    \includegraphics[width=0.9\linewidth, height=7.5cm]{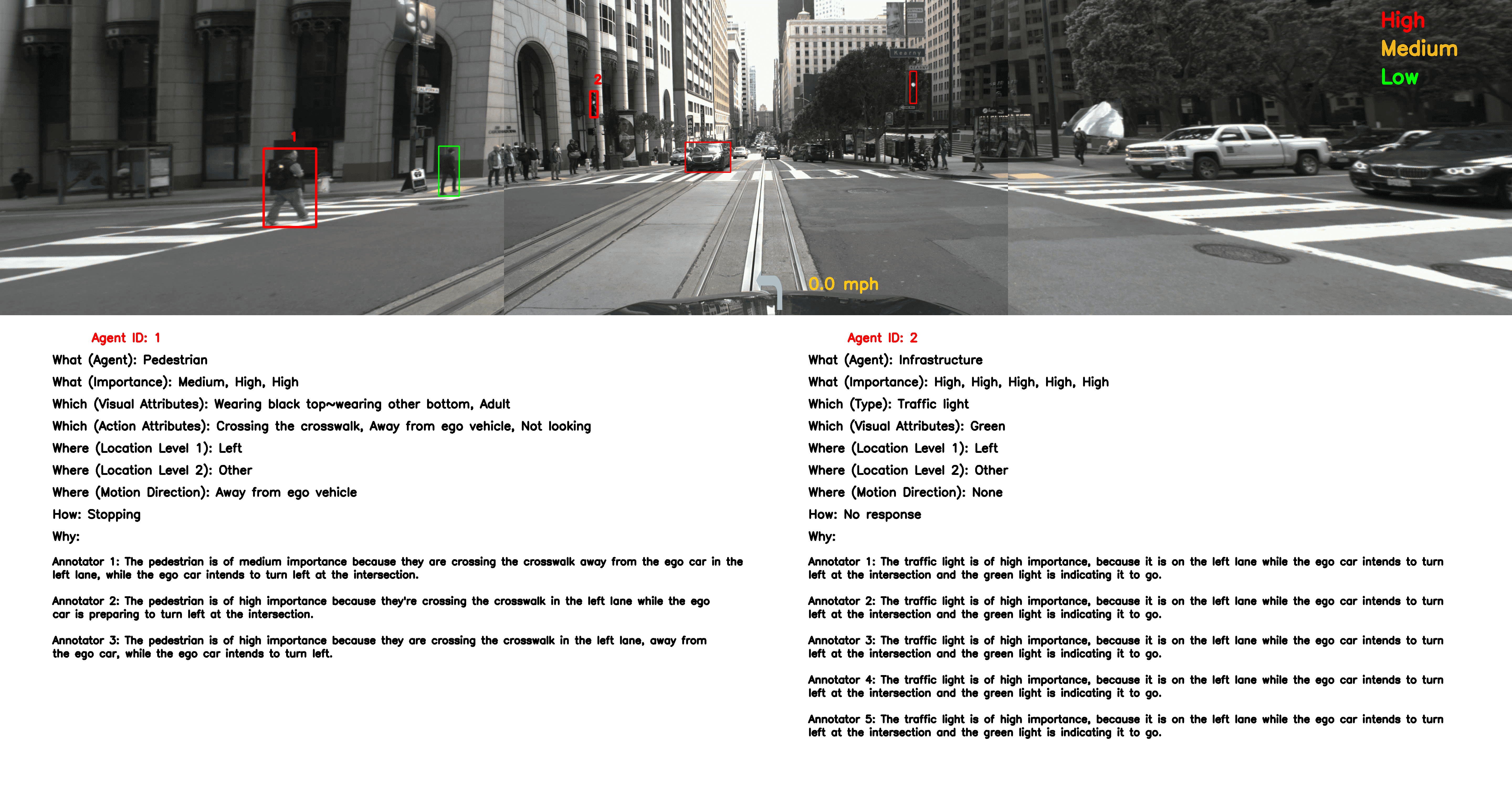}
    \caption{}
     \label{fig:left_intent_3}
  \end{subfigure}
    \caption{{\textbf{Example scenarios of Rank2Tell with ego intention to turn left at the intersection}}} 
    \label{fig:example_scenarios_left}
\end{figure*}

  

We show several frames from different scenarios of our introduced dataset-Rank2Tell, in Fig. \ref{fig:example_scenarios_straight}, \ref{fig:example_scenarios_right}, and \ref{fig:example_scenarios_left} with intentions of going straight, turning left, or turning right at the intersection. These frames are extracted from the scenarios, while the ego vehicle is entering the intersection. As shown, the ego vehicle's intention (straight, right or left) and speed (in mph) are displayed on the frames. Moreover, Fig. \ref{fig:example_scenarios_straight}, \ref{fig:example_scenarios_right}, and \ref{fig:example_scenarios_left} show the annotators responses to various close and open-ended visual questions in the form of 4W+1H (What, Which, Where, Why, How). Further, we take the mode of the various importance levels from different annotators as the final importance level. An agent with $High$, $Medium$, and $Low$ as the final importance level is shown in $red$, $yellow$ and $green$ color. 

The video corresponding to these scenarios is in the attached supplementary video. The video provides a comprehensive overview of the different scenarios and the corresponding annotations across time frames, offering a valuable insights in understanding the development of importance levels in complex driving situations.

In Fig. \ref{fig:straight_intent_1}, we observe that the ego vehicle's intention is to go straight. All five annotators deem the pedestrian crossing the cross-walk and the stop sign to be important, with the majority of them assigning a high importance level. The visual attributes for the pedestrian, such as color of the dress (top, bottom) and age, are captured in the \textit{Which} category. Similarly, the color of the stop sign is annotated as the visual attribute for infrastructure. The \textit{Where (action attributes)} section shows the pedestrian's action in the current frame. Furthermore, the \textit{Where (location level-1, location level-2, motion direction)} category captures the pedestrian's location and motion direction according to the annotation schema. The \textit{How} section records the ego vehicle's response to the important agent. As the pedestrian crosses the cross-walk, the ego vehicle decides to yield. Since the annotators can view the previous four seconds of the video, they can understand the yielding behavior based on the change in speed. Additionally, the \textit{Why} category shows the diverse reasons provided by the annotators based on their selected importance level for the pedestrian. The annotations effectively capture the responses of \textit{3W + 1H (What, Which, Where, How)} to generate a caption for the response to the \textit{Why}. These annotations demonstrate the usefulness of these structured visual question answering methods in perceiving risks and reasoning about them.

Moreover, in Fig. \ref{fig:straight_intent_2}, we see that the vehicle exiting the intersection from the right side is considered of medium importance by the majority of annotators, as it could influence the ego's behavior if it decides to make a lane change. Similarly, in Fig. \ref{fig:straight_intent_3}, the vehicle exiting the intersection from the right side is considered of high importance by the majority of annotators, as the front end of the vehicle is dangerously close to the ego vehicle, and the latter is slowing down to avoid a collision.

In Fig. \ref{fig:right_intent_1}, where ego intends to turn right at the intersection, the annotators assigned $High$ and $Medium$ importance levels to the pedestrians crossing the cross-walk on the right of the ego-vehicle. Specifically, for the pedestrian with ID 1 is marked as important by 4 out of 5 annotators, while 
that with ID 2 is marked as important by only 1 annotator. The difference in perceived importance levels comes from various factors such as annotator's driving experience, age, gender, etc. The captions provided by annotators for agent with ID 1, show consistency in reasoning. They all capture the location, motion direction and the intention of the ego vehicle to reason about the importance of an agent. In Fig. \ref{fig:right_intent_2}, the majority of annotators mark all pedestrians as $High$ important. This could be because of the close proximity of the pedestrians, as compared to the pedestrians in Fig. \ref{fig:right_intent_1}. Further, Fig \ref{fig:right_intent_3} shows that 3 and 5 annotators marked the truck and the crossing pedestrian of $High$ importance, because they directly influence the ego's decision. However, 2 annotators didn't mark the pedestrian to be of any importance. This discrepancy in annotations could be due to experienced drivers tending to consider only objects in close proximity to be of high importance, which is the truck in this case.

In Fig. \ref{fig:left_intent_1}, the Sedan at the intersection, and the bicycle turning left at the intersection are marked as $High$ importance by 2 and 3 annotators respectively. In Fig. \ref{fig:left_intent_2} and Fig \ref{fig:left_intent_3}, the agents towards the left of the intersection are marked as important by various annotators.  

These annotations demonstrate the effectiveness of the Rank2Tell dataset in capturing diverse responses of agents and infrastructure with various intentions of ego vehicles. The annotations show how annotators identified important visual attributes, actions, and locations of the agents in the scene, and how they reasoned about the importance of these agents based on their potential impact on the ego vehicle. The annotations also show that different annotators may assign different levels of importance to the same agents, based on their personal experience and perception of risk. This highlights the need for a diverse dataset like Rank2Tell to capture a range of perspectives and improve the robustness of machine learning models for autonomous driving.

These examples show that estimating importance level of various agents in the scene depends on various factors, and we aim to provide comprehensive annotations of those factors to address the problem of important agents classification and natural language explaination.



\subsection{Dataset Analysis}
\subsubsection{Diverse Annotations}

The diversity in Rank2Tell is one of its strengths, as it presents broad spectrum of diverse annotations. An agent’s importance is annotated with the prior knowledge of ego vehicle’s intention (left, right, straight) at the intersection. For instance, if the ego car intends to turn right while an agent is located on the left of the intersection, it is highly probable that none of the anno-
tators will deem it as important. In case a more conservative
annotator identifies it as important, users have the flexibility
to disregard the outliers to calculate the groundtruth impor-
tance, guided by various heuristics tailored to their specific
use case of the dataset, since we plan to release all 5 anno-
tators data with the release of the dataset. It also provides
users the flexibility to consolidate 4 levels of importance to
2 levels(important, non-important), based on their use case.

\subsubsection{Training, Validation and Testing Data}
The Rank2Tell dataset consists of total of 116 recordings, and we split the dataset into thress subsets of 70, 23, 23 scenarios for training, validation and testing data, respectively. We use the same split across all the baselines experiments, and report the results on the validation data.

Table \ref{tab:agents_dist} shows the distribution of various agent types and their importance levels in the three data subsets. The table shows the number of scenarios, and total frames in the 3 data splits. Additionally, it shows the distribution of 4 different agents types (vehicles, bicyclists, pedestrians, infrastructure) and their annotated importance levels by different annotators. For instance, it shows that $5283$ vehicles are being annotated as of $High$ importance in the training data, while $2604$ and $1818$ vehicles are being annotated as of $Medium$ and $Low$ importance in the validation and testing data. This comprises $40.1\%$, $46.7\%$, and $38.7\%$ of all important vehicles in the training, validation and testing data. Further, vehicles with importance level as $Medium$ consists of $38.7\%$, $37.5\%$ and $40.5\%$ of important vehicles in training, validation and testing data. And the vehicles with importance level as $Low$ consists of $21.1\%$, $15.7\%$ and $20.7\%$ of all important vehicles in training, validation and testing data. We observe comparable distribution across other agent types, as shown in Table \ref{tab:agents_dist}. This distribution indicates that the data splits consists of comparable number of data samples for different importance levels. We see similar distribution across other agents types- bicyclists, pedestrians and infrastructures. 

Table \ref{tab:vqa_dist} shows the distribution of visual question answering for 3 data splits. The dataset provides $29311$ importance levels responses from various annotators across $9305$ unique important agents in the training data.
It is important to note that the visual and action attributes of the agents consist of multiple sub-attributes associated with features such as color, age, and communicative and visual aspects. Therefore, the numerical values associated with these attributes are significantly higher than the number of unique important objects.

\begin{table*}[!tbp]
\centering
\setlength\tabcolsep{0pt} 

\begin{subtable}[t]{\textwidth}

\begin{tabular*}{\textwidth}{@{\extracolsep{\fill}}l|c|c|c|c|c|c|c|c|c|c|c|c|c|c} & \multicolumn{10}{c}{Agents Importance} \\ 
\hline
\hline

\multirow{2}{*}{Data} &
       \thead{$\#$\\Scenarios}  &   \thead{$\#$\\Frames}& 
      \multicolumn{3}{c|}{Vehicles} &
      \multicolumn{3}{c|}{Bicyclists} & 
      \multicolumn{3}{c|}{Pedestrians} &
      \multicolumn{3}{c}{Infrastructures}  \\   
        {} & & & {High} & {Medium} & {Low} & {High} & {Medium} & {Low} & {High} & {Medium} & {Low} & {High} & {Medium} & {Low}  \\
      \hline

      {Training} & 70 & 2364 & {5283} & {5101} & {2782} & {249} & {186} & {179} & {2997} & {1833} & {1540} & {8308} & {499} & {354} \\

      {Validation} & 23 & 784 & {2604} & {2091} & {881} & {180} & {161} & {87} & {682} & {460} & {640} & {2556} & {110} & {47} \\

      {Testing} & 23 & 685 & {1818} & {1905} & {973} & {62} & {145} & {122} & {416} & {437} & {218} & {2356} & {379} & {60} \\

      \hline

\end{tabular*}
\caption{Agents Importance level distribution for train, validation and test data split}
\label{tab:agents_dist}
\end{subtable}

\quad

\begin{subtable}[t]{\textwidth}

\centering
\begin{tabular*}{\textwidth}{@{\extracolsep{\fill}}l|c|c|c|c|c|c|c|c|c|c|c} & \multicolumn{10}{c}{Visual Question Answering} \\ 
\hline
\hline

\multirow{2}{*}{Data} & 
    \multicolumn{1}{c|}{\thead{All\\Agents}}&
      \multicolumn{2}{c|}{What} &
      \multicolumn{3}{c|}{Which} & 
      \multicolumn{3}{c|}{Where} &
      \multicolumn{1}{c|}{How}& 
      \multicolumn{1}{c}{Why}\\   
        & \thead{Important and\\ Non-Important} & \thead{Important\\Agents} & {Importance} & {Types} & \thead{Visual\\Attributes}  & \thead{Action\\Attributes} & \thead{Location\\Level-1} &  \thead{Location\\Level-2} & \thead{Motion\\Direction} & \thead{Ego's\\Response} & {Caption}  \\
      \hline
        {Training} &  {35839} & {9305} &  {29311} & {22327} & {36295} & {47284} & {29311} & {29311} & {20150} & {29311} & {29311}  \\
        
        {Validation} & {11970} & {3285} &{10499} & {8289} & {12709} & {17782} & {10499} & {10499} & {7786} & {10499} & {10499}  \\
        
        {Testing} & {9719} & {2782} & {8891} & {7491} & {10291} & {13592} & {8891} & {8891} & {6096} & {8891} & {8891}  \\

      \hline

\end{tabular*}
\caption{Distribution of Annotators' VQA responses for train, validation and test data split}
\label{tab:vqa_dist}
\end{subtable}
\caption{Annotators' data distribution }
\label{tab:ann_dist}
\end{table*}

\subsubsection{Annotators' data distributions}
\begin{figure*}[!htp]
\centering
\subfloat{
\includegraphics[width = 2\columnwidth,height=4.0cm]{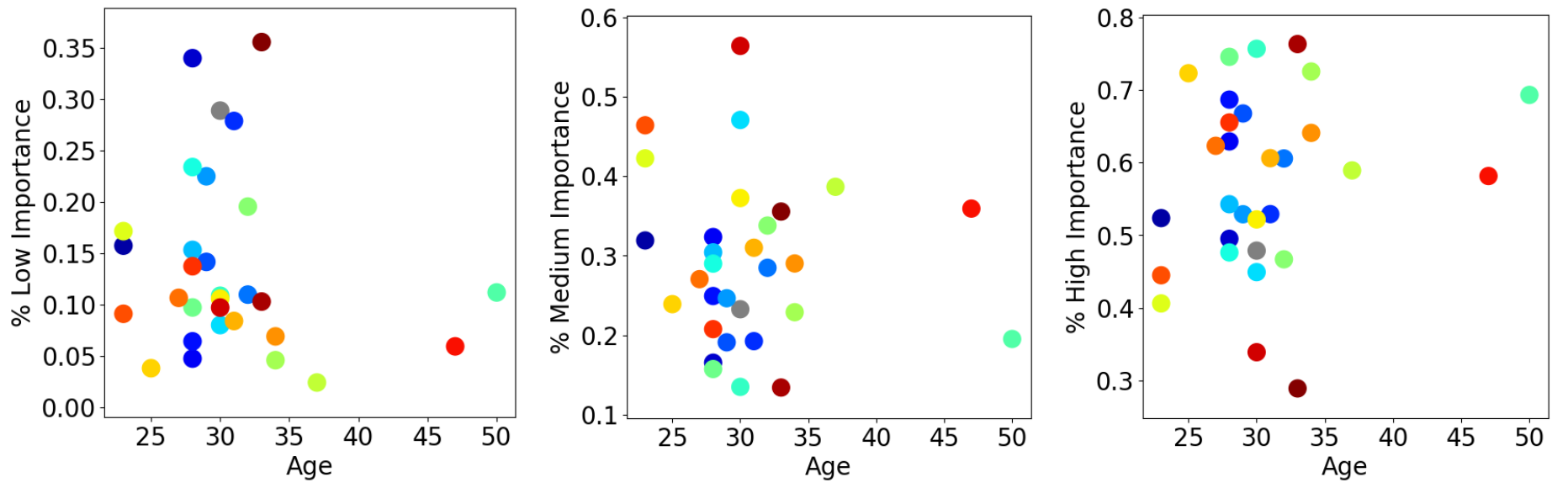}} \\
\subfloat{
\includegraphics[width = 2\columnwidth,height=4.0cm]{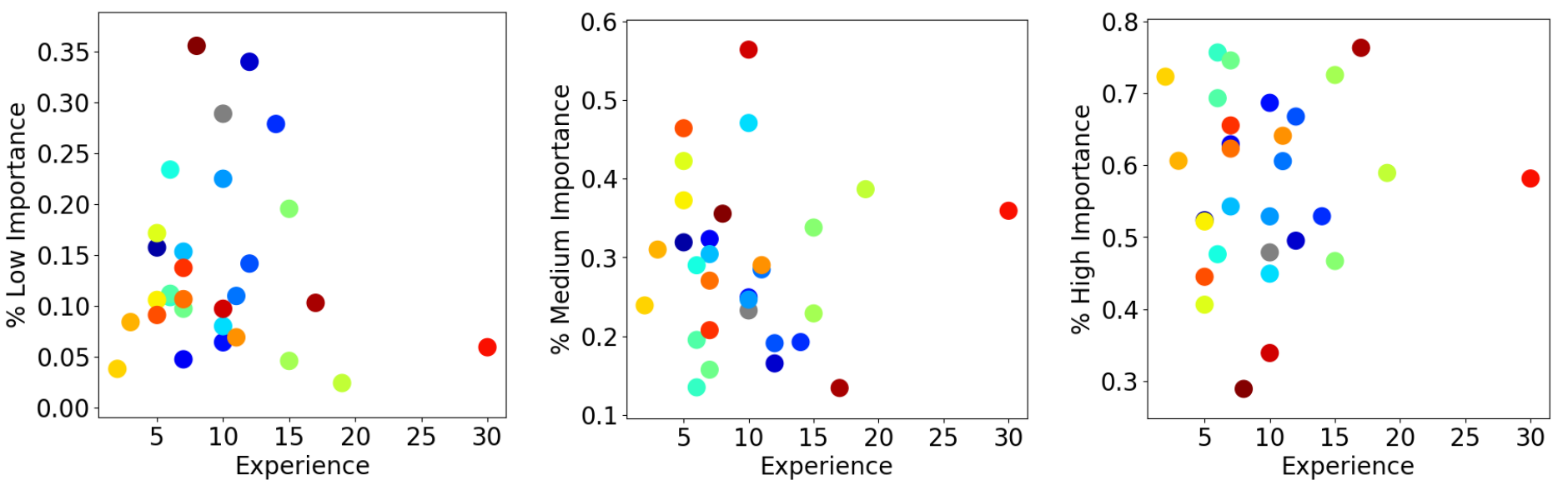}} \\
\subfloat{
\includegraphics[width = 2\columnwidth,height=4.0cm]{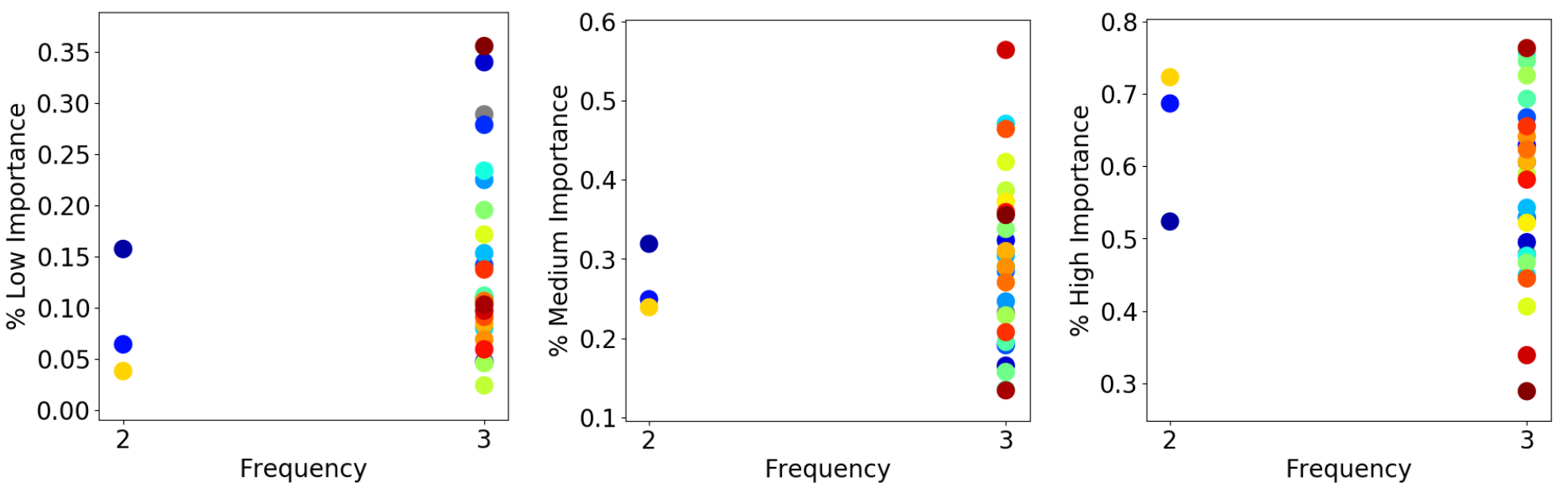}} \\
\subfloat{
\includegraphics[width = 2\columnwidth,height=4.0cm]{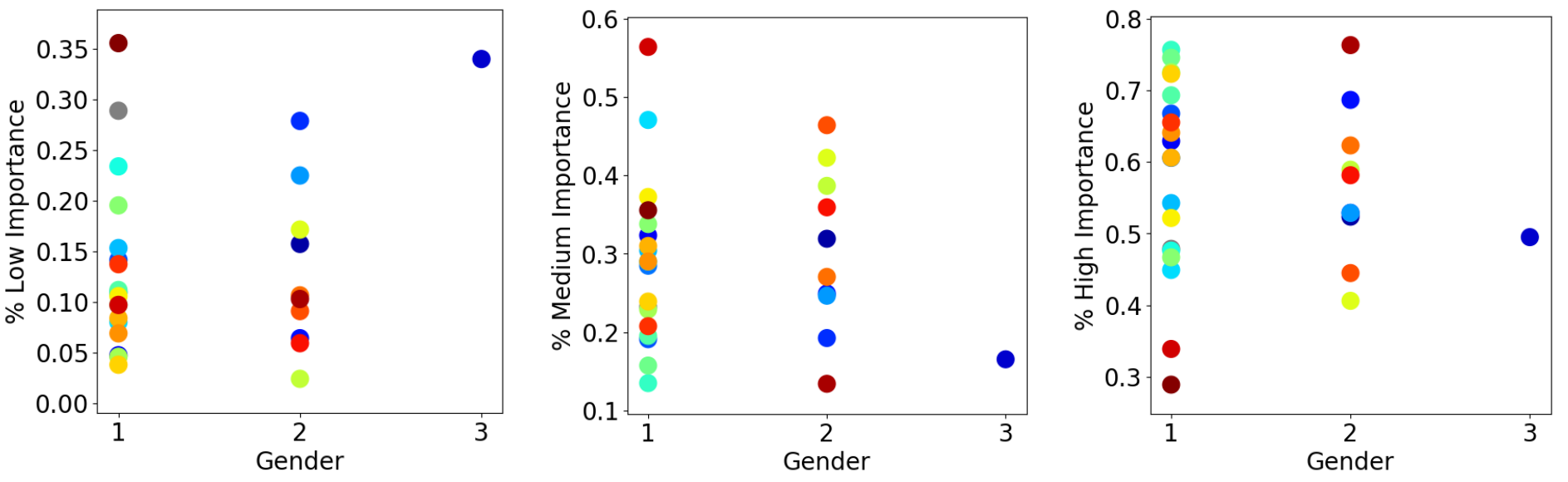}} \\
\subfloat{
\includegraphics[width = 2\columnwidth,height=4.0cm]{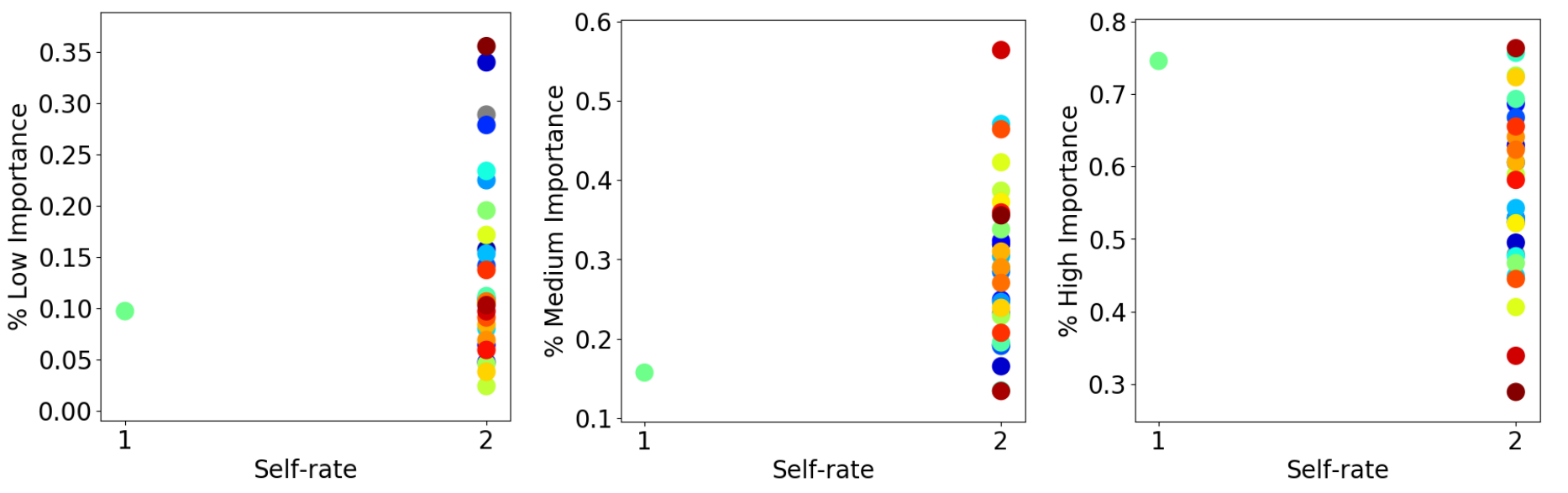}}
\caption{Relationship between importance level (grouped by columns) and annotator personal information (grouped by rows).
Each annotator has been assigned a unique color, and is represented in each figure by a dot. From top row: (1) age in years, (2) driving experience
in years, (3) frequency of driving, either 1-rarely, less than once a month, 2-occasionally, about once a
week, 3-frequently, more than three times a week, (4) gender 1-male, 2-female, 3-not sure (5) rating of driving skill, 1-intermediate,
2-advanced.}
\label{ann_stats}
\end{figure*}

Figure \ref{ann_stats} presents the relationship between different importance levels and annotators' personal information. The dataset employs 43 unique annotators; however, due to privacy concerns, not all annotators share their personal information. Thus, we have complete information for 35 annotators, while 7 annotators only provided their gender and age, and 1 annotator only provided their gender.

For the 116 scenarios in our dataset, the asked questions resulted in a total of 2900 data points, out of which we were able to collect 2585 (89.1\% of the total).
As shown in Figure \ref{ann_stats}, there is a strong correlation between the number of years of experience and importance ranking annotations. Additionally, we observe a similar trend with respect to age.


\section{Implementation Details}

\subsection{Groundtruth Importance}
 Table 2 in the main paper highlights the consistency and quality of importance level across 5 annotators. This is calculated using mode of all 5 annotators importance levels as the final importance of an agent in the scene. This includes importance as well as non-importance levels. Nonetheless, when transitioning to practical use cases of not underestimating an agent's importance, we use groundtruth as the mode of only importance annotations, i.e if 2 of out 5 annotators deem an agent as important, we take the mode of only 2 annotators, and not all 5. In cases of multiple modes, we take the highest importance level as the groundtruth importance. Table \ref{tab:consistency_imp2} shows the consistency using this method for estimating the groundtruth importance.  A consistency of $20\%$, $40\%$, $60\%$, $80\%$, $100\%$ indicates that all five annotators deem an object as important, while only 1, 2, 3, 4, and 5 annotators, respectively, provided the same importance level as the mode. A consistency of $33.33\%$ and $66.66\%$ indicates that three annotators deem an object as important, while only 1 and 2 annotators, respectively, provided the same importance level as the mode. Lastly, the consistency of $50\%$ indicates that if 2 annotators deem an object as important, both provided different importance, and if 4 annotators deem an object as importance, half (2) annotators provided the same importance level, and $75\%$ indicates that 4 annotators deem an object as important, while only 3 provided the same importance level as the mode. It shows that for data instances with High, Low and Non-Important as groundtruth importance, $89.3\%$, $83.20\%$ and $100\%$ of samples exhibited more than $60\%$ consistency. This suggests that annotators had a high level of agreement on objects marked as High, Low and Non-important. However, for data instances with Medium as the majority importance, only $75.65\%$ of samples had more than $60\%$ consistency. This could be attributed to the ambiguity between medium and high importance, and medium and low importance. Nonetheless, consistency score doesn’t demonstrate the quality of the dataset, but shows the agreement/disagreement among annotators with a groundtruth as final importance.




    
    






\begin{table}[!h]
\centering
\small

\begin{tabular}{c|ccc}
                      & \multicolumn{3}{c}{Agents Importance} \\ \hline
\thead{Consistency} & \thead{Low} & \thead{Medium} & \thead{High}\\ 

    \hline
  
    \hline
    20 &  0 & 0 & 0\\
    \hline
    33.33 & 2.12 & 1.03 & 0.51 \\
    \hline
    40 &  1.44 & 5.36 & 2.69 \\
    \hline
    50 &  13.21 & 17.93 & 7.46 \\
    \hline
    60 & 3.85 & 13.72 & 9.18 \\
    \hline
    66.66 &  8.00 & 8.36 & 4.44 \\
    \hline
    75 &  3.79 & 5.31 & 4.76 \\
    \hline
    80 &   0.96 & 7.71 & 13.73\\
    \hline
    100 &  66.60 & 40.55 & 57.19\\
    \hline

    $>=$60 &   83.20 & 75.65 & 89.3\\
    \hline

\end{tabular}
\caption{Inter-Annotator consistency of importance annotation based on mode of only importance annotations}
\label{tab:consistency_imp2}
\end{table}

\subsection{Joint Model}

\subsection{2D Deep-Feature extraction}
The 2D deep feature extractor uses a sequence of RGB images, depth images, semantic maps, ego vehicle's inertial data, and 2D bounding box of each objects in the scene. The depth images are obtained by projecting the pointcloud to the stitched camera view, and the segmentation maps are obtained using DeepLabv3 on the stitched RGB images. Visual feature extractor adopts ResNet101 pretrained on ImageNet dataset as the backbone. It takes in RGB images to generate object level features and global features of all frames, which are then fed into Sequence Encoder I to obtain image features $v_{{2D}_{(j,A)}}$ and $v_{global,A}$. Depth feature extractor adopts ResNet18 trained from scratch as a backbone, and takes in concatenated depth and semantic segmentation maps to generate object level and features and global features of all frames, which are fed into Sequence Encoder III to obtain final features, $v_{{2D}_{(j,DS)}}$ and $v_{global,DS}$. To extract the features of each object from  $v_{{2D}_{(j,A)}}$ and $v_{{2D}_{(j,DS)}}$, a ROIAlign pooling layer is added before feeding to sequence encoder. 
The final visual feature of each object, and global context information is obtained by concatenation as $v_{{2D}_{j}} = [v_{{2D}_{(j,A)}}, v_{{2D}_{(j,DS)}}]$, $v_{global} = [v_{global,A}, v_{global,DS}]$. The ego vehicle features $v_{ego}$ are obtained using Ego State Feature Encoder, and the bounding box features of other agents $v_{{2D}_{(j,B)}}$, are obtained using Sequence encoder II, as shown in Fig. 5 in the main paper.

\subsection{3D Deep-Feature extraction}
We represent the pointcloud data as $\mathcal{P} = {(p_i, f_i)} \in \mathcal{R}^{N_p \times 6}$, where $p_i \in  \mathcal{R}^3, i= 1,2, ...., N_p$ are the coordinates, and  $f_i \in \mathcal{R}^{3}$ are the additional features, representing radius, confidence and curvature. The point cloud features consist of $<x, y, z, radius, confidence, curvature>$ data points. Additionally, we utilize the 3D bounding boxes of agents in the scene, obtained from LIDAR annotations, as well as the instance and semantic labels associated with each point cloud. The output of voting module is a set of point clusters $ \mathcal{C} \in \mathcal{R}^{M \times 128}$, where M is the upper bound of the number of proposals. We assume that the bounding boxes of objects (i.e., vehicles, infrastructure, pedestrians, and bicyclists) can be obtained using off-the-shelf detection and tracking system in advance. Therefore, we leverage the groundtruth 3D bounding boxes, and groundtruth objects coordinates information for the votenet model to obtain bounding boxes features $v_{{3D}_{j,P}}$. 


The final object level features are obtained by concatenating the corresponding 2D features and 3D features: 
$v_{j} = [v_{{2D}_{j}}, v_{{3D}_{j}}]$    


\subsection{Relational Feature Extraction}

The importance level ranking and reasoning of an object in a scene are influenced by the state and appearance of other objects in the scene. Thus, to capture the mutual influence and relations among objects, we use a graph neural networks that models objects as nodes and their relations as edges in a relational graph module. The module takes in the concatenated object features and extracts both object features $v_{j}$ and relational features $e_{i,j}$ between objects. 

By applying a message passing mechanism over the graph, the nodes can exchange information with their neighbors, allowing them to update their features based on the features of their neighbors. To model the relational (edges) features, the module considers only the $K$ nearest objects surrounding the target object to limit the computation complexity.

\begin{equation}
    \begin{aligned}
v \rightarrow e : e_{ij}= f^1_e([v_{i}, v_{j}]),
    \end{aligned}
\end{equation}

\begin{equation}
    \begin{aligned}
e \rightarrow \bar v : v_i = f_v(\Sigma_{i \neq j}e_{(i, j)}),
    \end{aligned}
\end{equation}

The final objects features are obtained by concatenating the graph nodes features with the object relations, global features, ego features, and the ego intentions $I_E$: $o_{j} = [v_{j}, \bar  v_{j}, v_{global}, v_{ego}, I_E]$

\subsection{Importance Classification}
The final object features is fed into the classifier to obtain its importance class (high, medium, low or binary). The classifier outputs the logits corresponding to different importance level as $y = [\hat {y_1}, ..., \hat {y_{n-1}}]$.

\subsection{Captioning Decoder}
The captioning module takes the object features $(o_j)$ to generate caption with one token at a time, using GRU. Similar to Scan2Cap \cite{chen2021scan2cap}, we add the relational features between object $j$ and its neighbor $k$ to the corresponding $o_j$ to obtain the final attention context feature set $V^r = {v^r_1, ..., v^r_j , ..., v^r_M}$, where  $v^r_j = o_j + \Sigma_{k=1}^M \mathcal{e}_{jk}$. The intermediate attention distribution over the context features are defined as:

\begin{equation}
    \alpha_t := softmax((\mathcal{V}^r W_v + \mathds{1}_h h_{t-1}^{T}W_h)W_a)\mathds{1}_a
\end{equation}

The attention module outputs the aggregated context vector  
$\hat{v_t} = \Sigma_{i=1}^M \mathcal{V}_i^r  \odot \alpha_{ti}$ to represent the attended object and corresponding inter-object relation. The language GRUs uses the $\hat{v_t}$ and hidden state $h_t^2$ to predict the token $y_t$ at current step $t$.

\subsection{Training}

The joint model predicts the importance of all agents within the scene - both important and not-important. However, the model solely generates captions of important agents. Subsequent to the Graph Neural Network (GNN) module, the agents deemed non-important are systematically filtered out. This filtration utilizes the ground truth importance level both during training and evaluation phases. Such an approach ensures a fair comparison with other captioning baselines, which have been modified similarly to exclude non-important agents based on the ground truth importance level.

The training objective combines the caption loss and the importance classification loss in the following manner: 

\begin{equation}
\mathcal{L} = \alpha \mathcal{L}_{imp} + \beta \mathcal{L}_{cap}
\end{equation}
where $\alpha$, $\beta$ are the weights for the individual loss terms.

TO enforce the model to reduce the instances of falsely underestimating an agent’s importance, we penalize the $L_{imp}$ corresponding to different groundtruth (GT) and predictions (P) for different importance levels of high (H), medium(M), low(L), and not-importance(NI) as follows:

\begin{align}
\mathcal{L}_{imp}  & = \Sigma_{i=1}^{N} \mathcal{L}_{i} \\
\mathcal{L}_{i} & =\begin{cases}
    \lambda_k \mathcal{L}_{i}, & \text{if $(GT - P) = k > 0$}.\\
    \mathcal{L}_{i} , & \text{otherwise}.
  \end{cases}
\end{align}
where $\mathcal{L}_{i}$ is the cross-entropy loss for each object $i$. 

For 4 classes, levels $0$, $1$ $2$ and $3$ represent $Non-Important$, $Low$, $Medium$ and $High$ importance, respectively. For binary classes, levels $0$ and $1$ represent $Non-Important$, and $Important$, respectively.
 
\begin{itemize}
 \item Predicted Importance is 1 level lower than groundtruth importance: If groundruth is $High$, prediction is $Medium$, or groundtruth is $Medium$, prediction is $Low$, or groundtruth is 
 $Low$, prediction is $Not-Important$ (for 4 classes), or groundtruth is $Important$, prediction is $Not-Important$ (for 2 classes), we weigh the loss by $\lambda_1$

  \item Predicted Importance is 2 level lower than groundtruth importance: If groundruth is $High$, prediction is $Low$, or groundtruth is $Medium$, prediction is $Not-Important$, we weigh the loss by $\lambda_2$

 \item Predicted Importance is 3 level lower than groundtruth importance: If groundruth is $High$, prediction is $Not-Important$, we weigh the loss by $\lambda_3$

 \item Predicted Importance is higher than groundtruth importance: groundruth is $Not-Important$, prediction is $High$, or $Medium$ or $Low$ importance, or groundruth is $Low$, prediction is $High$, or $Medium$, groundtruth is $Medium$, prediction is $High$ (for 4 classes), or groundtruth is $Not-Important$, prediction is $Important$ (for 2 classes), we weigh the loss by $1$
\end{itemize}

For joint model, we use the following values of parameters:
 $\lambda_1$ = $\lambda_2$ = $\lambda_3$ = 1, and $\alpha$ = $\beta$ = 1


\subsection{Importance level classification}
For a fair comparison with the joint model, all three baselines~\cite{li2022important, zhang2020interaction, gao2019goal} use the same feature extractor ResNet101~\cite{he2016deep} pre-trained on the ImageNet dataset with Feature Pyramid Networks~\cite{lin2017feature} on top. During training, we adopt stochastic gradient descent with ADAM optimizer to learn the network parameters.  The model is trained for 100 epochs using an initial learning rate of 0.0001 and a learning rate decay of 10. All feature layers are jointly updated during training. For consistency, we maintain a fixed input resolution of $5760\times 1200$ and use a batch size of 32 for all experiments.

\begin{table}[!tbp]
\centering
\begin{tabular}{c|ccc} 
Method       & F1 (I) & F1 (NI) & Accuracy \\ \hline
Ours w/o actions & \textbf{78.44} & 92.97 &  89.39\\
Ours w/ actions & 78.27 & \textbf{93.01} &  \textbf{89.42}\\ 

\end{tabular}
\caption{Quantitative Evaluation comparing the F1 scores for 2 importance levels of our joint model, with and without augmented action features. I: IMPORTANT, NI: NON-IMPORTANT}
\label{table:binary}
\end{table}

\begin{table}[!tbp]
\label{binary_class}
\centering
\begin{tabular}{c|cccc} 
Method    &  C & B-4 & M & R \\ \hline
Ours w/o actions&  100.15 &  45.83 & 36.21 & 68.56 \\
Ours w/ actions&  \textbf{107.14}  &  \textbf{47.45}  & \textbf{36.75} & \textbf{69.41}  \\


\end{tabular}
\caption{Quantitative Evaluation comparing the performance for captions predictions of our joint model, with and without augmented action features. C: CIDER, B-4: Bleu-4, M: Meteor, R: Rouge}
\vspace{-0.3cm}
\label{table:caption}
\end{table}

\subsection{Captioning Baselines}
For Scan2Cap \cite{chen2021scan2cap} baseline, we use 3D pointcloud data with $<x, y, z, radius, confidence, curvature>$ as the pointcloud features. The point cloud features consist of $<x, y, z, radius, confidence, curvature>$ data points. Additionally, we utilize the 3D bounding boxes of agents in the scene, obtained from LIDAR annotations, as well as the instance and semantic labels associated with each point cloud. For a fair comparison with our proposed model, we assume that the bounding boxes of objects, such as vehicles, pedestrians, and bicyclists, can be obtained using off-the-shelf detection and tracking systems in advance. Consequently, we leverage the ground truth 3D bounding boxes and object instance information to train the votenet module to learn the bounding box features. This is a modified version of the Scan2Cap model that does not perform object detection and instead relies on the ground truth bounding boxes information to learn the bounding box features. We limit the maximum number of objects in a particular frame to 64. However, it is important to note that our current work does not include infrastructure due to the lack of available 3D pointcloud features for traffic lights, stop signs, speed limit signs, etc. To ensure a fair comparison, we do not include infrastructure for the S$\&$T \cite{vinyals2015show} baseline as well. Nevertheless, we plan to incorporate infrastructure in our future extensions of this work.





\subsection{Integrating Action Attributes}

To demonstrate the usefulness of these annotations, we conduct an ablation study where we integrate action attributes (Which) with object features in a simple way within the joint model. We concatenate one hot vector of action attributes with object features before the GNN, and the results in Table \ref{table:binary} and Table~\ref{table:caption} show superior performance of the model as compared to those without action.

{\small
\bibliographystyle{ieee_fullname}
\bibliography{egbib}
}
